\documentclass{article} 
\usepackage{ACL2023}

\usepackage{amsmath,amsfonts,bm}

\def\eqref#1{equation~\ref{#1}}

\def\1{\bm{1}}

\DeclareMathAlphabet{\mathsfit}{\encodingdefault}{\sfdefault}{m}{sl}
\SetMathAlphabet{\mathsfit}{bold}{\encodingdefault}{\sfdefault}{bx}{n}

\usepackage[utf8]{inputenc} 
\usepackage[T1]{fontenc}    
\usepackage{hyperref}       
\usepackage{url}            
\usepackage{booktabs}       
\usepackage{amsfonts}       
\usepackage{nicefrac}       
\usepackage{microtype}      
\usepackage{xcolor}         
\usepackage{graphicx}
\usepackage{times}
\usepackage{latexsym}
\usepackage{multirow}
\usepackage{balance}
\usepackage{amsmath}
\usepackage{bbm}
\usepackage{paralist}
\usepackage{makecell}
\usepackage{amssymb}
\usepackage{subcaption}
\usepackage{caption}
\usepackage{xspace}
\usepackage{multicol}
\usepackage{transparent}
\usepackage{array}
\usepackage{bm}
\usepackage{tabularx}
\usepackage{floatrow, makecell}
\usepackage{wrapfig, lipsum,booktabs}
\usepackage{float}
\usepackage{enumitem}
\usepackage{pifont}%
\newcommand{\cmark}{\ding{51}}%
\newcommand{\xmark}{\ding{55}}%

\definecolor{midnightgreen}{rgb}{0.0, 0.33, 0.33}

\definecolor{red}{rgb}{1.0, 0.0, 0.0}

\definecolor{darkmagenta}{RGB}{139, 0, 139}

\newcommand{\bs}[1]{\boldsymbol{#1}}
\newcommand{\model}{MoMA\xspace} 

\title{Augmenting Zero-Shot Dense Retrievers with Plug-in Mixture-of-Memories}

\author{
   Suyu Ge$^{1}$\thanks{\hspace{5pt}Work partly done during Suyu's internship at Microsoft.} , Chenyan Xiong$^2$, Corby Rosset$^2$, Arnold Overwijk$^2$, Jiawei Han$^1$, Paul Bennett$^2$\\
 	 \textsuperscript{1} University of Illinois Urbana-Champaign \quad  \textsuperscript{2} Microsoft Research \\ 
 	 \texttt{\{suyuge2,hanj\}@illinois.edu}\\
 	\texttt{\{chenyan.xiong,corbyrosset,arnold.overwijk,paul.n.bennett\}@microsoft.com}
  }   

\begin{document}

\maketitle

\begin{abstract}
In this paper we improve the zero-shot generalization ability of language models via Mixture-Of-Memory Augmentation (MoMA), a mechanism that retrieves augmentation documents from multiple information corpora (``external memories''), with the option to ``plug in'' new memory at inference time.
We develop a joint learning mechanism that trains the augmentation component with latent labels derived from the end retrieval task, paired with hard negatives from the memory mixture.
We instantiate the model in a zero-shot dense retrieval setting by augmenting a strong T5-based retriever with MoMA.
Our model, \model{}, obtains strong zero-shot retrieval accuracy on the eighteen tasks included in the standard BEIR benchmark.
It outperforms systems that seek generalization from increased model parameters and computation steps.
Our analysis further illustrates the necessity of augmenting with mixture-of-memory for robust generalization, the benefits of augmentation learning, and how \model{} utilizes the plug-in memory at inference time without changing its parameters.
We plan to open source our code.
\end{abstract}
\section{Introduction}

Scaling up language models---with more parameters, compute, and annotation data---improves model generalization ability on downstream applications~\citep{raffel2019t5, brown2020language, smith2022using}, but with diminishing return: \textit{linear}  improvements on downstream metrics often require \textit{exponentially} more parameters and computing cost~\citep{kaplan2020scaling, hoffmann2022training}. 
Hence, scaling pretrained language models in this way is economically unsustainable~\citep{strubell2020energy, bender2021dangers, zhang2022opt}.

Retrieval augmented language models provide a promising alternative. They allow language models to efficiently access vast resources from an external corpus~\citep{guu2020realm, borgeaud2022improving} that serves as a kind of ``memory'' they can refer to when making predictions, alleviating the need to memorize as much information in their own network parameters~\citep{roberts2020much}.
This open-book approach helps language models to better generalize on token prediction tasks and machine translation~\citep{khandelwal2019generalization, borgeaud2022improving}, and tasks which already involve a first-stage retrieval component, e.g., OpenQA~\citep{ borgeaud2022improving, izacard2022few}.
Existing retrieval augmentation methods usually stick to one single retrieval corpus throughout training and inference so that the retrieval component can be indirectly guided by the supervision from end tasks.

In this paper we improve the zero-shot generalization ability of language models using ``mixture-of-memory'' (MoMA), a new retrieval augmentation mechanism.
Instead of a single corpus, MoMA retrieves  documents from a ``mixture'' of multiple external corpora and enjoys the merits of a larger and more comprehensive source of knowledge.
This mechanism also allows removing and/or ``plugging-in'' new corpora during inference time, when more information from the target task is revealed, or as an additional way for users to control the model.
Specifically, we apply MoMA on the zero-shot dense retrieval task, which is the foundation of many important real-world applications~\citep{thakur2021beir, kim2022applications} and also the retrieval component of recent retrieval augmented language models~\citep{guu2020realm, izacard2022few}.
However, it is not trivial to guide a retrieval model to leverage multiple corpora.
We need to jointly train the augmentation component and dense retriever using supervised relevance signals and self-mined hard negatives.




We instantiate MoMA with a T5 encoder-decoder model~\citep{ni2022sentence} and apply it to the dense retrieval task~\citep{karpukhin2020dense}. 
Our end task retriever uses a set of augmenting documents from the mixture-of-memories to enhance its representation of the query with important context; the retriever then uses the enhanced query representation to retrieve a final candidate set. 
At inference time, we plug in the target task's corpus to the memory mixture to introduce in-domain context information, without updating any parameter.

We experimented on eighteen zero-shot dense retrieval tasks included in BEIR~\citep{thakur2021beir}, the standard ZeroDR benchmark.
The results demonstrate the improved zero-shot ability of \model{}.
When paired with the ANCE~\citep{xiong2020approximate} training framework on a T5 model, it outperforms counterparts without the MoMA augmentation component, as well as recent state-of-the-art dense retrieval systems of the same scale, by large margins.
To validate its effectiveness when paired with advanced models, we further instantiate \model{} with a contrastively pretrained T5 model.
\model{} then achieves comparable or even stronger performance to ZeroDR systems with larger model scales and heavier computation costs.



Our analysis reveals that large and diverse corpora in the memory leads to the best performance; 
while only using a single corpus during training does not improve performance on unseen target tasks.
The learning of augmentation component is also important for \model{} to utilize the diverse information from the mixture.
Our analysis and case studies illustrate how \model{} leverages the plug-in memory at testing time to enrich its query representations with in-domain information that was not available in training.

\section{Related Work}
\subsection{Retrieval Augmentation}
Recent research has explored two common ways to construct the external memory in retrieval-augmented language models.
The first is to retrieve similar tokens for language models to copy from when predicting the next token~\citep{khandelwal2019generalization, zhong2022training}.
The second is to retrieve the related documents (text sequences) from an in-domain corpus as additional input~\citep{guu2020realm, borgeaud2022improving}.
Our work falls into this category as document-based models better align with knowledge-intensive tasks~\citep{petroni2020kilt}, such as retrieval and OpenQA~\citep{chen2017reading}.

Learning to retrieve useful documents to augment the language model is a challenging task, since human annotations on the usefulness of augmentation documents are costly and seldom available.
The most straightforward way is to use representations from raw pretrained language models to find documents similar to the task input, i.e., as unsupervised dense retrieval~\citep{guu2020realm, borgeaud2022improving}.
Adapting dense retrieval models trained for relevance matching is another common choice~\citep{izacard2020leveraging, lewis2020retrieval, yu2021improving}.
A more formal solution is to jointly learn the augmentation components end-to-end using supervision from the final task,
for example, treating the augmentation as latent variables and applying EM~\citep{zhao2021distantly}, or distilling the augmentation component from feedback of the final model~\citep{izacard2020distilling}.
In a parallel work, \citet{izacard2022few} found the most effective one is attention distillation method (ADist), which trains the augmentation component using soft labels derived from the end model's attention on augmentation documents.

The motivation for query augmentation coincides with the query expansion methods in the traditional IR community, whereby the user’s original query is augmented by new features with similar meanings~\citep{carpineto2012survey}.
As feature selection usually requires additional semantic analysis, the efficiency and usability of traditional query expansion methods remain limited when faced with a new domain.
To overcome this, recent work relies on dense retrieval results to expand the query~\citep{yu2021improving}.
The retrieved relevant documents serve as pseudo relevance feedback signals for the model, which are concatenated with the original query as the augmented model input.
Our work augments queries with feedback from multiple corpora and learns to select important augmentation documents automatically.




\subsection{Zero-shot Dense Retrieval}
Dense retrieval models trained on a resource rich source tasks, e.g., web search, usually do not perform as well when zero-shot transferred to other domains ~\citep{beir}.
This is concerning since many important real-world scenarios do not have the luxury of web corpus training signals and must rely on near zero-shot transfer, e.g., the medical domains~\citep{kim2022applications}.
\citet{xin2021zero} analyzed the challenge of shifting between training and testing domains, and leveraged domain-invariant learning to mitigate the gap. 
Another common approach is to first generate domain-specific pseudo labels for each task, and then use them to train dense retriever~\citep{beir, wang2021gpl}.
Additionally, continuous pretraining the language model also improves its generalization ability in ZeroDR~\citep{izacard2021towards, gao2022cocondenser,yu2022coco}.
Following works~\citep{izacard2021towards,yu2022coco} further contrastively pretrained the retriever on source or target corpus with a sentence matching loss.
Other methods seek better generalization ability in ZeroDR from various resources, for example, combining with sparse retrieval to introduce exact match signals~\citep{formal2021splade}, using multiple vectors per documents for term-level matching~\citep{khattab2020colbert}, or scaling up the retrieval model using larger language models~\citep{gtr, cpt}.

\section{Method}

In this section we first describe our Mixture-of-Memory Augmentation. Then we discuss how it is jointly learned with the end system and enables plug-in memory at inference time.

\subsection{Mixture-of-Memory Augmentation}
Before going to the details of MoMA, we first recap some preliminaries in ZeroDR.

\textbf{Preliminaries.} The dense retrieval (DR) task aims to find relevant documents $d$ from a corpus $C$ for the given query $q$ by representing them in a shared embedding space. Specifically, the retrieval score in DR is often calculated as:
\begin{align}
    f(q, d) &= \bs{q} \cdot \bs{d};  \bs{q} = g(q); \bs{d} = g(d). 
\end{align}
It uses dot product as the scoring function to match the embeddings $\bs{q}$ and $\bs{d}$, which is known to support efficient nearest neighbor search (ANN)~\citep{johnson2019billion}. 
A pretrained language model is often the encoder of choice $g()$. 
We use the ST5-EncDec variant of Sentence-T5~\citep{ni2022sentence}:
\begin{align}
    g(x) &= \text{Dec}(\text{Enc}(x)), \label{eqn:st5}
\end{align}
which feeds in the text sequence (prepended by a special [CLS] tokens) to the encoder of T5, $\text{Enc}()$, and uses the output representation of the [CLS] token from the decoder, $\text{Dec}()$, as the text representation. This naturally leverages the attention from decoder to encoder at all Transformer layers~\citep{raffel2019t5}, as a fine-grained information gathering mechanism.

The \textit{training} of dense retrieval systems often applies standard ranking loss and pairs the relevant documents $d^+ \in D^+$ for each query q with hard negatives $d^- \in D^-$:
\begin{align}
    \mathcal{L} = \sum_{q} \sum_{d^+ \in D^+} &\sum_{d^- \in D^-} l(f(q, d^+), f(q,d^-)); \nonumber \\
    D^- &\sim \text{ANN}_{f(q, \circ)}^C \setminus D^+. \label{eqn:ance}
\end{align}
Eqn.~\ref{eqn:ance} uses ANCE hard negatives, which are the top-retrieved documents from $C$ using the retriever itself~\citep{xiong2020approximate}. The loss function $l()$ can be any standard ranking loss such as cross entropy.
A ZeroDR model is trained on $q^s$ and documents $d^s \in C^s$ from a \textit{source task}, often web search, and tested on \textit{target} tasks $q^t$ and $C^t$; supervision signals are only present from the source.

\textbf{Mixture-of-Memory Augmentation.} 
The key idea of (document-based) retrieval augmented language models is to enrich the representation $g(q)$ with additional contextual input for the model, i.e., augmentation documents $d^a$ retrieved from an external memory $\mathcal{M}$. Instead of using a single document corpus, MoMA uses multiple corpora to provide richer and more diverse external resources for augmentation. For example, $\mathcal{M}$ can be composed by the source corpus $C^s$, a general encyclopedia, a domain specific knowledge graph, etc. Then we can retrieve the augmentation documents $D^a$ :
\begin{align}
    D^a &= \text{ANN}_{f^a(x,\circ)}^\mathcal{M}; \text{ } \mathcal{M} = \{C_1,...,C_M\}. \label{eqn:moma}
\end{align}
This augmentation component uses another dense retriever $f^a()$ (also a Sentence T5 model), with parameters distinct from those in $g()$. Note that instead of retrieving $D^a$ separately from $M$ different ANN memory sources and merging results, Eqn.~\ref{eqn:moma}  combines them into one ANN index. This requires the augmentation component $f^a()$ to be flexible enough handle various corpora in the mixture.

\begin{figure}
\centering
\includegraphics[width=\textwidth]{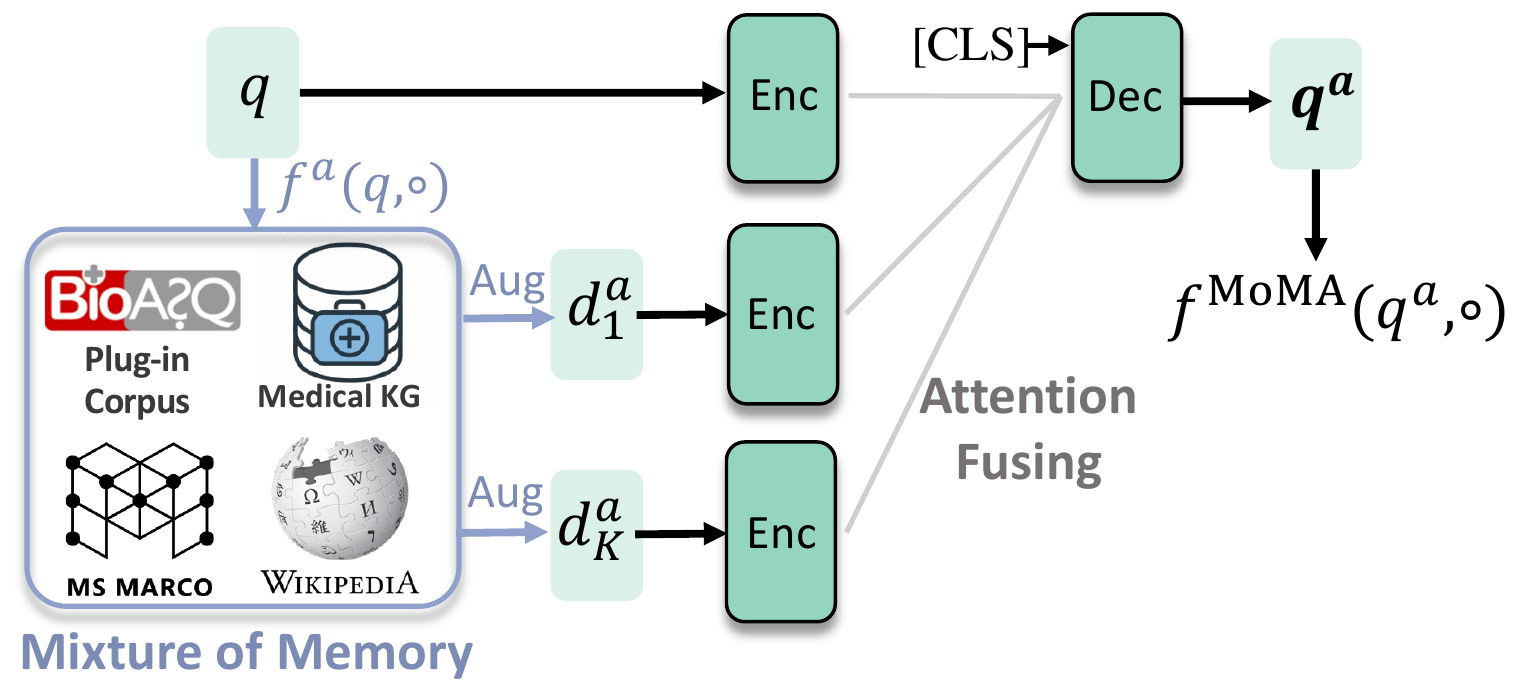}
\caption{Illustraion of the Mixture-of-Memory Augmentation. \label{fig:model}}
\vspace{-0.3cm}
\end{figure}

Using the encoder-decoder architecture for $g()$ in Eqn.~\ref{eqn:st5} enables a simple extension to incorporate the augmentation documents using the fusion-in-decoder (FiD) mechanism~\citep{izacard2020leveraging}:
\begin{align}
    g^\text{MoMA}(q) &= \text{Dec}(\text{Enc}(q),\text{Enc}(d_1^a),...,\text{Enc}(d_K^a)); \nonumber \\
    & \text{  } D^a = \{d^a_1,...,d^a_K\}.    \label{eqn:moma_g}
\end{align}
It feeds in the $K$ augmentation documents separately to the T5 encoder of $g()$. Then it fuses the encoded documents together with $\text{Enc}(q)$ using one decoder that attends to all encoded vectors, as illustrated in Figure~\ref{fig:model}. 

The FiD approach in Eqn~\ref{eqn:moma_g} is a nice balance of efficiency and capacity when modeling multiple text sequences~\citep{izacard2020leveraging}. It is more efficient than concatenating all text pieces together, while also remaining expressive enough to model the nuances from many sequences.
~\citep{izacard2020distilling, izacard2022few}.

When instantiating MoMA in the dense retrieval setting, we focus on augmenting the query representation $\bs{q}$, as queries are often short, ambiguous, and benefit more from additional contextual information~\citep{lavrenko2017relevance, yu2021improving}. This leads to the following definition of \model{}:
\begin{align}
     f^\text{MoMA}(q, d) = & \bs{q}^a \cdot \bs{d}; \nonumber \\    
     \bs{q}^a = g^\text{MoMA}(q), & \bs{d} = g(d),   \label{eq:moma_dr}
\end{align}
using the construction of $g^\text{MoMA}()$ in Eqn.~\ref{eqn:moma_g} upon the augmentation documents defined in Eqn.~\ref{eqn:moma}.

\subsection{Joint Learning in \model{} and Inference with Plug In Memory} 

\model{} has two sets of parameters to learn, in the main model $f^\text{MoMA}()$ and the augmentation component $f^a()$. Both have their own T5 encoder-decoder parameters.
The two components are bridged by the augmentation documents, which are retrieved by $f^a()$ from  $\mathcal{M}$ and used by $f^\text{MoMA}()$ to produce query representation $\bs{q}^a$.

\textbf{Main Model Learning.} Given the relevance labels from the source task and an augmentation model, training  $f^\text{MoMA}()$ is straightforward. We can use the standard dense retrieval training to finetune the enriched query encoder $g^\text{MoMA}()$ and the document encoder $g()$:
\begin{align}
    \mathcal{L}^\text{\tiny{MoMA}} &= \sum_{q^s} \sum_{d^+} \sum_{d^-} l(f^\text{\tiny{MoMA}}(q^s, d^+), f^\text{\tiny{MoMA}}(q^s,d^-));  \nonumber \\
    &d^+ \in D^{s+}, d^- \in D^{s-}  \label{eqn:l_moma} \\ 
    &D^{s-} \sim \text{ANN}_{f^\text{MoMA}(q^s, \circ)}^{C^s} \setminus D^{s+}.
    \label{eqn:ance_moma}
\end{align}
The training signals come from the source task, including $q^s$, its relevant documents $D^{s+}$, and ANCE hard negatives $D^{s-}$ retrieved from the source corpus $C^s$.

\textbf{Augmentation Learning.} Training $f^a()$ is challenging as it is hard to  label whether an augmentation document is useful.
Propagating gradients from the final loss to $f^a()$ is also prohibitive as the retrieval operation in Eqn.~\ref{eqn:moma} is discrete. 
Fortunately, recent research found the attention scores from the FiD decoder to each encoded inputs (Eqn.~\ref{eqn:moma_g}) are good approximations to the usefulness of augmentation documents~\citep{izacard2020distilling}:
\begin{align}
    \text{FidAtt}(d^a_i) &= \sum_{\text{layers}} \sum_{\text{positions}} \sum_{\text{heads}} \text{Att}_{\text{Dec} \rightarrow \text{Enc}}(g^\text{MoMA}(d^a_i)). \label{eqn:fidatt}
\end{align}
It sums the attentions from $g^\text{MoMA}()$'s special token at the decoder's [CLS] position over all layers, input positions, and attention heads.
Ideally, higher $\text{FidAtt}()$ is assigned to $d^a_i$ that provides useful contextual information.

Previously, FidAtt scores are often used as soft labels for the augmentation model~\citep{izacard2020distilling, izacard2022few}. Doing so with memory mixtures is risky as it is too sparse and overfits memory resource that appears earlier in the training, which are the only ones available for the decoder to attend on.
To improve the learning robustness, we introduce ANCE-style hard negative mining to train the augmentation component as well.

First, we formulate the positive set of augmentation documents as:
\begin{align}
    D^{a+} &= D^{s+} \cup \text{Top-N}_{\text{FidAtt}(d^a_i), D^a}.
    \label{equ:topn}
\end{align}
which combines relevant documents $D^{s+}$ and the augmenting ones that received N-highest attention scores from $g^\text{MoMA}()$. 
Then we pair them with hard negatives to formulate the training of $f^a()$ as:
\begin{align}
    \mathcal{L}^a &= \sum_{q^s} \sum_{d^+ \in D^{a+}} \sum_{d^- \in D^{a-}} l(f^a(q^s, d^+), f^a(q^s,d^-)); \label{eqn:l_a} \\
    &D^{a-} \sim \text{ANN}_{f^a(q^s, \circ)}^{\mathcal{M}} \setminus D^{a+}. 
\end{align}
Notice the negatives for $f^a()$ have comprehensive coverage from multiple corpora.

\textbf{Iterative Training.} The learning of $f^{\text{MoMA}}()$ and $f^a()$ is an iterative process that fits naturally into the training procedure of dense retrieval training with hard negatives. We follow the standard iterations in ANCE and construct the $t$-th training episode of \model{}:
\begin{enumerate}

    \item Construct hard negatives $D^{s-}$ via Eqn.~\ref{eqn:ance_moma} using weights $f_{t-1}^\text{MoMA}()$ from the last episode;
    \item Retrieve augmentation $D^a$ via Eqn.~\ref{eqn:moma} using weights $f_{t-1}^a()$ from the last episode;
    \item Train $f_{t}^\text{MoMA}()$ as Eqn.~\ref{eqn:l_moma};
    \item Formulate new positive augmentation documents $D^{a+}$, using updated attention scores from $f_{t}^\text{MoMA}()$, and mine negative augmentation documents $D^{a-}$ using $f_{t-1}^a()$;
    \item Train $f_{t}^a()$ following Eqn.~\ref{eqn:l_a}.
\end{enumerate}
Both $f_0^\text{MoMA}()$ and $f_0^a()$ can be initialized with a BM25 warmed-up T5 retriever. Steps 1 and 3 above are inherited from standard dense retrieval training.  The rest are introduced by MoMA. 
The additional computation in the training side mainly resides updating the index for the memory mixture, a standard cost in retrieval-augmented language models~\citep{guu2020realm, izacard2022few}.

\textbf{Zero-Shot Retrieval with Plug in Memories.} 
To perform zero-shot retrieval on unseen tasks, \model{} first retrieves augmented documents using $f^a()$ from $\mathcal{M}$ for the target query $q^t$, and retrieves target documents $d^t \in C^t$ with the augmented model $f^\text{MoMA}()$ without changing any model parameters. MoMA allows $f^a()$ to attend over the target corpus as well if it is plugged in: $\mathcal{M} = \mathcal{M} \cup C^t \setminus C^s$, which conveys in-domain information. 
The augmenting corpus can also be engineered by users manually to inject their preference or domain knowledge, e.g., as ``memory engineering''. 
In this work we focus on swapping out the source corpus for the target corpus; we leave other explorations for future work.

\section{Experimental Methodologies}
\label{sec:exp_setting}

\textbf{Datasets.}
We choose the MS MARCO passage dataset~\citep{msmarco} as the source domain dataset, whereas the target domains are from the 18 datasets in BEIR~\citep{beir} benchmark, which include including biomedical, scientific and financial texts.
More details can be found in Appendix~\ref{appx:datasets}.
The evaluation metric NDCG@10 is the same with BEIR benchmark, which measures Normalized Discounted Cumulative Gain~\citep{ndcg} of top 10 prediction. 
The higher NDCG@10 value indicates better performance.

\textbf{Augmenting Corpora.} 
During training, the mixture-of-memory is composed of source training corpus (MARCO), Wikipedia and a medical knowledge graph.
We use the Wikipedia chunk prepossessed by~\cite{karpukhin2020dense} without further processing\footnote{https://huggingface.co/datasets/wiki\_dpr}.
The medical knowledge graph is extracted from the Medical Subject Headings (MeSH)\footnote{https://www.ncbi.nlm.nih.gov/mesh/}, an open-source database for indexing and cataloging of biomedical and health-related information.
Since it is hierarchical in structure, we linearize it by concatenating spans with text information.
During testing, we directly replace MARCO with the corresponding document sets from BEIR.
Each task from BEIR is augmented independently.
More dataset and preprocessing details can be found in Appendix~\ref{appx:datasets}.

\begin{table*}[h]
  \small
  \centering
  \resizebox{\linewidth}{!}{%
\begin{tabular}{@{}l|ccccccccccccccc@{}}
\toprule 
 &\multirow{2}{*}{BM25}&\multirow{2}{*}{DPR}&\multirow{2}{*}{ANCE}&\multirow{2}{*}{{T5-ANCE}}&{\multirow{2}{*}{coCondenser}}&{\multirow{2}{*}{GenQ$^\dagger$}}& {\multirow{2}{*}{ColBERT}}& {\multirow{2}{*}{Contriever}} &{\multirow{2}{*}{GTR{\tiny{base}}$^*$}}&{\multirow{2}{*}{GTR{\tiny{large}}$^*\ddagger$}}&{\multirow{2}{*}{\makecell[c]{\textbf{MoMA} \\ \textbf{(T5-ANCE)}}}}&{\multirow{2}{*}{\makecell[c]{\textbf{MoMA} \\ \textbf{(COCO)}}}}  \\ 
 \\
\hline
 \textbf{Parameters\#}&{---}& 110M & 110M & 110M*2 & 110M & 66M*18 & 110M & 110M& 110M & 335M & 110M*2 & 110M*2 \\ 
\hline
TREC-COVID    & {0.656}     & {0.575} & {0.654} & {0.653} & {0.715} & {0.619} & {0.677} & {0.596} & {0.539} & {0.557} & \textbf{0.762} & {0.761} \\
BioASQ        & {0.465}     & {0.232} & {0.306} & {0.322} & {0.318} & {0.398} & \textbf{0.474} & { --- } & {0.271} & {0.320} & {0.372} & {0.371} \\
NFCorpus      & {0.325}     & {0.210} & {0.237} & {0.275} & {0.307} & {0.319} & {0.305} & {0.328} & {0.308} & {0.329} & {0.307} & \textbf{0.333} \\
NQ            & {0.329}     & {0.398} & {0.446} & {0.452} & {0.494} & {0.358} & {0.524} & {0.498} & {0.495} & \textbf{0.547} & {0.490} & {0.544} \\
HotpotQA      & {0.603}     & {0.371} & {0.456} & {0.487} & {0.566} & {0.534} & {0.593} & \textbf{0.638} & {0.535} & {0.579} & {0.539} & {0.589} \\
FiQA-2018     & {0.236}     & {0.274} & {0.295} & {0.294} & {0.285} & {0.308} & {0.317} & {0.329} & {0.349} & \textbf{0.424} & {0.320} & {0.329} \\
Signal-1M     & \textbf{0.330}     & {0.238} & {0.249} & {0.246} & {0.274} & {0.281} & {0.274} & { --- } & {0.261} & {0.265} & {0.258} & {0.264} \\
TREC-NEWS     & {0.398}     & {0.366} & {0.382} & {0.379} & {0.389} & {0.396} & {0.393} & { --- } & {0.337} & {0.343} & {0.413} & \textbf{0.453} \\
Robust04      & {0.408}     & {0.344} & {0.392} & {0.412} & {0.399} & {0.362} & {0.391} & { --- } & {0.437} & {0.470} & {0.469} & \textbf{0.475} \\
ArguAna       & {0.414}     & {0.414} & {0.415} & {0.415} & {0.411} & {0.493} & {0.233} & {0.446} & {0.511} & \textbf{0.525} & {0.438} & {0.463} \\
Touché-2020   & \textbf{0.367}     & {0.208} & {0.240} & {0.312} & {0.190} & {0.182} & {0.202} & {0.230} & {0.205} & {0.219} & {0.271} & {0.299} \\
Quora         & {0.789}     & {0.842} & {0.852} & {0.836} & {0.863} & {0.830} & {0.854} & {0.865} & {0.881} & \textbf{0.890} & {0.847} & {0.843} \\
DBPedia-entity& {0.313}     & {0.236} & {0.281} & {0.290} & {0.356} & {0.328} & {0.392} & \textbf{0.413} & {0.347} & {0.391} & {0.347} & {0.383} \\
SCIDOCS       & {0.158}     & {0.107} & {0.122} & {0.115} & {0.140} & {0.143} & {0.145} & \textbf{0.165} & {0.149} & {0.158} & {0.143} & {0.145} \\
Fever         & {0.753}     & {0.589} & {0.669} & {0.655} & {0.678} & {0.669} & \textbf{0.771} & {0.758} & {0.660} & {0.712} & {0.723} & {0.745} \\
Climate-Fever & {0.213}     & {0.176} & {0.198} & {0.194} & {0.184} & {0.175} & {0.184} & {0.237} & {0.241} & \textbf{0.262} & {0.235} & {0.233} \\
SciFact       & {0.665}     & {0.475} & {0.507} & {0.566} & {0.600} & {0.644} & {0.671} & \textbf{0.677} & {0.600} & {0.639} & {0.632} & {0.630} \\
CQADupStack   & {0.299}     & {0.281} & {0.296} & {0.283} & {0.330} & {0.347} & {0.350} & {0.345} & {0.357} & \textbf{0.384} & {0.283} & {0.294} \\ 
\hline
Contriever Sub Avg &0.437&0.368&0.408&0.416& {0.438} & {0.425} & {0.445} & {0.466} & {0.442} & {0.471} & {0.453} & \textbf{0.471} \\
{Avg}       & {0.428}     & {0.352} & {0.391} & {0.399} & {0.417} & {0.410} & {0.431} & { --- } & {0.416} & {0.444} & {0.436} & \textbf{0.453} \\
\bottomrule
\end{tabular}}
\caption{
NDCG@10 on the BEIR benchmark. 
We also include an averaged score on datasets used by Contriever for a fair comparison.
The best result each task is marked bold.
An $^*$ denotes unfair comparison, as NQ is used in training for GTR. 
$\dagger$: GenQ generated pseudo labels to train an independent model for each task.\
$\ddagger$: Larger models
}
\label{tab:all_ndcg}
\vspace{-0.2cm}
\end{table*}

\begin{table}[h]
    \centering
    \resizebox{0.95\linewidth}{!}{
    \begin{tabular}{l|ccc}
    \toprule 
    \textbf{Model}& \textbf{Pretraining Corpus} & \textbf{Batch Siz}e& \textbf{Training Steps} \\
    \hline
    MoMA (T5-ANCE)& 0 & 0 & 0 \\
    \hline
    MoMA (COCO) & MARCO & 128 & 50k \\
    \hline
    GTR & NQ, CQA & 2048 & 800k \\
    \hline
    Contriever & CCNet & 2048 & 500k\\
     & Wikipedia & 2048 & 200k\\
    \bottomrule 
    \end{tabular}}
    \caption{Computational analysis in the  pretraining stage of different models.}
    \label{tab:pretrain_cost}
    \vspace{-0.3cm}
\end{table}

\textbf{Baselines and Model Choices.}
We compare our \model with standard sparse and dense retrieval models on BEIR.
We also compare \model with advanced approaches that are specifically designed for zero-shot generalization.
They involve techniques that are not directly comparable with this paper, including pretraining on extra data, in-domain continuous pretraining, and generating target pairs using another pretrained generative model.
Besides, some baselines use larger scale language model as their backbone.
We list the details of baselines in Appendix~\ref{appx:baseline}.

As a plug-in-and-play method, MoMA can be combined with other techniques.
We initiate MoMA on two versions of T5 model checkpoints. 
The primitive \textbf{MoMA (T5-ANCE)} is built on the original T5 model checkpoint.
By comparing it with T5-ANCE, we can clearly observe the performance gain brought by MoMA.
To demonstrate it can integrate techniques from other models to achieve higher performances, we apply MoMA with a better pretrained T5-based model.
Following previous work~\citep{gao2022cocondenser,yu2022coco}, we continuously trained the T5 model on the MARCO corpus using a sentence-level contrastive loss, combined with the original masked language modeling loss.
We then performed the same MoMA training on top of the continuously pretrained T5 checkpoint and denoted it as \textbf{MoMA (COCO)}.
Both \textbf{MoMA (T5-ANCE)} and \textbf{MoMA (COCO)} are trained iteratively with ANCE-style~\citep{xiong2020approximate} hard negatives, the only difference is the initialized model start point.
We compare their pretraining details with other models in Table~\ref{tab:pretrain_cost}.
Unlike previous work~\citep{yu2022coco}, we did not include target datasets and augmenting corpora in the COCO pretraining stage.
Since MARCO contains only 0.5M documents, it adds fewer computational overhead compared to other methods listed in the table, e.g., Contriever.

\textbf{Implementation Details.}
For \model{}, we use the T5-base~\citep{raffel2019t5} architecture  (12-layer Transformer, 768 hidden size) by directly loading the checkpoint from HuggingFace\footnote{https://huggingface.co/t5-base}.
To warm up the language model for dense retrieval, we followed \cite{xiong2020approximate} to further train it using BM25 negatives for 10 epochs. 
After warming up, we jointly trained the two components for three episodes, each episode including three training epochs.
After three joint episodes, the end retriever reaches the best performance on MSMARCO, so we select this checkpoint for evaluation.
The ratio between positive and hard negative pairs is 1:7 for both models.
The main hyperparameters in {\model} include the total number of grounding documents $K$ and the attention threshold number N in Equation~\ref{equ:topn}. 
We directly set $K$=10 and N=5 without any parameter tuning.
More details on hyperparameters and experimental settings can be found in Appendix~\ref{app:para}.
\section{Evaluation Results}

Our experiments evaluate the zero-shot ability of \model{}, its performance with different memory sources, the influence of memory mixture learning, and the benefits of plug-in memory.

\begin{table}[t]
    \centering
    \resizebox{0.95\linewidth}{!}{
    \begin{tabular}{l|cc}
    \toprule 
    \textbf{Operation} & \textbf{Offline} & \textbf{Online} \\
    \hline
    BM25 Index Build & 1.8h & --- \\
    BM25 Retrieval Per Query & --- & 43ms \\
    \hline
    \textbf{MoMA Inference} & & \\
    Encoding of Corpus/Per Doc &1.5h/4.5ms &---\\
    Query Encoding & --- & 55ms\\
    ANN Retrieval (batched q) & --- & 9ms\\
    Dense Retrieval Total & --- & 64ms\\
    \hline
    \textbf{MoMA Training} & & \\
    Encoding of Corpus/Per Doc & 1.5h/4.5ms &---\\
    ANN Index Build & 10s &---\\
    Neg Construction Per Batch (32 queries) & 45ms &---\\
    Back Propagation Per Batch (32 queries)& 330ms &---\\
    \bottomrule 
    \end{tabular}}
    \caption{Efficiency of MoMA search and training.}
    \label{tab:efficiency}
    \vspace{-0.2cm}
\end{table}

\subsection{Zero-Shot Retrieval Accuracy and Efficiency}
The retrieval accuracy of \model{} and baselines are listed in Table~\ref{tab:all_ndcg}. 
Besides baselines of similar parameter count, we also include larger models (GTR{\tiny{large}}) or those using multiple vectors per document (ColBERT).
{MoMA (COCO)} shows the strongest zero-shot accuracy against previous state-of-the-art methods that do continuous contrastive pretraining (coCondenser), generate pseudo labels (GenQ), or consume additional training signals in both continuous pretraining and finetuning phrases (GTR$_\text{base}$). 
{MoMA (T5-ANCE)} also achieved nearly comparable zero-shot accuracy against larger models like GTR$_\text{large}$, and ColBERT, which scales up the number of vectors per documents (one per token).
This confirms that retrieval-augmentation provides another path to improve language models' generalization ability besides scaling up. 
{MoMA (T5-ANCE)} also outperforms T5-ANCE, which {MoMA (T5-ANCE)} uses as a subroutine for retrieval augmentation, on all but one retrieval task, showing the robustly improved generalization ability from plug-in mixture of memory.

We evaluate the efficiency of MoMA in two stages: offline model training and online inference.
In offline training from Table~\ref{tab:pretrain_cost}, MoMA (T5-ANCE) is \textbf{significantly cheaper} than other methods as we do not require pretraining on large external corpora, which saves hundreds of hours training time.
MoMA (COCO) additionally pretrain on MARCO for 50k steps, which is far fewer than the other compared methods. 
In online inference, similar with other retrieval enhanced language models, MoMA imposes a necessary cost of retrieval augmented model upon the baseline T5-ANCE.
We further provide detailed efficiency analysis on MoMA in Table~\ref{tab:efficiency}.
The online latency is measured on one query and 100 retrieved documents. 
Due to the query augmentation, query encoding is more costly and takes about 55ms per query. 
Even with the augmentation cost, the full dense retrieval total online inference cost is 64ms, only slightly above the BM25 retrieval latency.
The ANN retrieval is very efficient, only takes 9ms. In addition, the complexity of ANN retrieval is sub-linear to the corpus size, in most ANN framework such as FAISS. Thus the extra round of ANN retrieval operation in MoMA is not the bottleneck even when the size of memory mixture scales up.

\subsection{Performance with Different Memories}
\label{sec:mem_ablation}

\begin{table*}[t]
  \centering
  \resizebox{0.8\linewidth}{!}{%
\begin{tabular}{@{}l|c|cccc|ccc@{}}
\toprule
 &{\textbf{No Memory}}&  \multicolumn{4}{c|}{\textbf{Single Memory}}& \multicolumn{3}{c}{\textbf{Memory Mixture}}\\ 
 \cline{2-9} 
 & T5-ANCE&{MARCO}&{Wiki}&{MeSH}&{Target}&{w/o Learning}&{w/o Target}&{Full} \\ 
\hline
TREC-COVID    &0.653 &0.576 &0.592 &0.669 &0.731 &\underline{0.759} &0.664 &\textbf{0.762} \\
BioASQ        &0.322 &0.247 &0.262 &0.219 &\underline{0.361} &0.359 &0.271 &\textbf{0.372}  \\
NFCorpus      &0.275 &0.295 &0.302 &0.282 &\textbf{0.319} &\underline{0.317} &0.301 &0.307 \\
NQ            &0.452 &0.472 &0.486 &0.393 &0.483 &\textbf{0.510} &0.484 &\underline{0.490} \\
HotpotQA      &0.487 &0.481 &0.519 &0.462 &\underline{0.538} &\textbf{0.539} &0.520 &\textbf{0.539} \\
FiQA-2018     &0.294 &0.296 &0.286 &0.280 &\textbf{0.320} &\underline{0.304} &0.285 &\textbf{0.320} \\
Signal-1M     &0.246 &0.239 &0.225 &0.238 &\underline{0.250} &0.248 &0.240 &\textbf{0.258} \\
TREC-NEWS     &0.379 &0.381 &0.391 &0.372 &\textbf{0.416} &0.410 &0.398 &\underline{0.413} \\
Robust04      &0.412 &0.435 &0.443 &0.428 &\textbf{0.483} &0.446 &0.452 &\underline{0.469} \\
ArguAna       &0.415 &\underline{0.439} &0.438 &\textbf{0.442} &\underline{0.439} &0.427 &0.438 &0.438 \\
Touché-2020   &\underline{0.312} &0.281 &0.281 &0.252 &\textbf{0.331} &0.275 &0.272 &0.271 \\
Quora         &\underline{0.836} &0.809 &0.798 &0.835 &0.781 &0.813 &0.812 &\textbf{0.847} \\
DBPedia-entity&0.290 &0.340 &0.341 &0.287 &0.335 &0.331 &\underline{0.342} &\textbf{0.347} \\
SCIDOCS       &0.115 &0.128 &0.121 &0.130 &\textbf{0.146} &0.134 &0.127 &\underline{0.143} \\
Fever         &0.655 &0.663 &\underline{0.735} &0.610 &0.694 &0.718 &\textbf{0.737} &0.723 \\
Climate-Fever &0.194 &0.231 &\underline{0.238} &0.231 &0.228 &0.222 &\textbf{0.240} &0.235 \\
SciFact       &0.566 &0.583 &0.587 &0.585 &\underline{0.624} &0.618 &0.598 &\textbf{0.632} \\
CQADupStack   &\textbf{0.283} &0.207 &0.218 &{0.203} &\textbf{0.283} &\underline{0.235} &0.215&\textbf{0.283}  \\
\hline
Avg           &0.399 &0.395 &0.403 &0.384 &\underline{0.431}&0.426 &0.411 &\textbf{0.436} \\
\bottomrule
\end{tabular}}
\caption{NDCG@10 of \model{} (T5-ANCE) under different memory compositions: no memory, single memory, and a mixture of memories. \textit{w/o Learning} uses the end retriever to select augmenting documents without use of an augmentation component. \textit{w/o Target} excludes the target from memory. 
\label{tab:memory_ablation}}
\end{table*}

Table~\ref{tab:memory_ablation} evaluates how \model{} behaves under different combinations of external memories. 
Compared with the MoMA (T5-ANCE), MoMA (COCO) may lean towards the MARCO corpus since it is continuously pretrained on it.
To avoid unfair comparison between MARCO and other corpora, we choose MoMA (T5-ANCE) as the \textit{Full} model version for ablation studies.
Unsurprisingly, using a single out-of-domain memory for retrieval augmentation does not help, for example, even though MARCO is the source domain corpus, solely grounding on it reduces zero-shot accuracy. 
MeSH as the sole augmenting corpus also lowers performance, even on some medical retrieval tasks such as BioASQ. 
Interestingly, when we expand the memory to include MARCO, Wiki, and MeSH, but keep the target corpus excluded (\textit{w/o Target}), \model{} exhibits better accuracy compared to the no-memory T5-ANCE. Our conclusion is that more memory sources achieves better generalization, especially when no target domain information is available.

In the \textit{Full} setting, the 3-memory mixture of MARCO, Wiki, and MeSH is jointly learned with final task at training time. At test time, MARCO  is swapped out for the target corpus. The \textit{Full} improves zero-shot accuracy over both the \textit{w/o Target} setting (where the target corpus is excluded at test time), and the \textit{w/o Learning} setting (wherein the augmentation component is not learned). 
As expected, plugging in the target corpus at test time is the most valuable source of generalization power. It is also the most realistic, as access to the target corpus may only be available at testing time. 

\subsection{Effect of Memory Mixture Learning}
\label{sec:exp_distill}
\begin{table*}[t!]
\small
\centering
\resizebox{0.9\columnwidth}{!}{
		\begin{tabular}{l|ccccccc}
		   \toprule
		    Distillation Method \quad \quad & TREC-COVID & BIOASQ & NFCorpus & NQ & HotpotQA & FEVER &\textbf{Avg}\\
		   \hline
		    \multicolumn{7}{l}{\textbf{Soft Attention Distill}}\\
		   \quad ADist~\citep{izacard2022few} &0.609&0.185&0.227&0.351&0.387&0.615&0.396 \\
		   \quad ADist + MSMARCO rel&0.664&0.220&0.255&0.397&0.394&0.624&0.426 \\
		  \textbf{w/o Distilling (Fixed)}&0.741&0.361&0.301&0.472&0.513&0.684&0.512 \\
     		 \textbf{\model{} (T5-ANCE)} &\textbf{0.762} &\textbf{0.372} &\textbf{0.307} &\textbf{0.490} &\textbf{0.539} &\textbf{0.723}&\textbf{0.532}\\
		   \hline 
		\end{tabular}
	}
	\caption{Zero-shot Performances of different distillation methods. We observe consistent trend on all BEIR datasets. 
	We present results on 6 representative datasets from Wikipedia or medical domains.\label{tab:distill}}
\vspace{-0.2cm}
\end{table*}

\begin{figure*}[t]
\begin{subfigure}[]{0.245\textwidth}
     \centering
     \includegraphics[width=\textwidth]{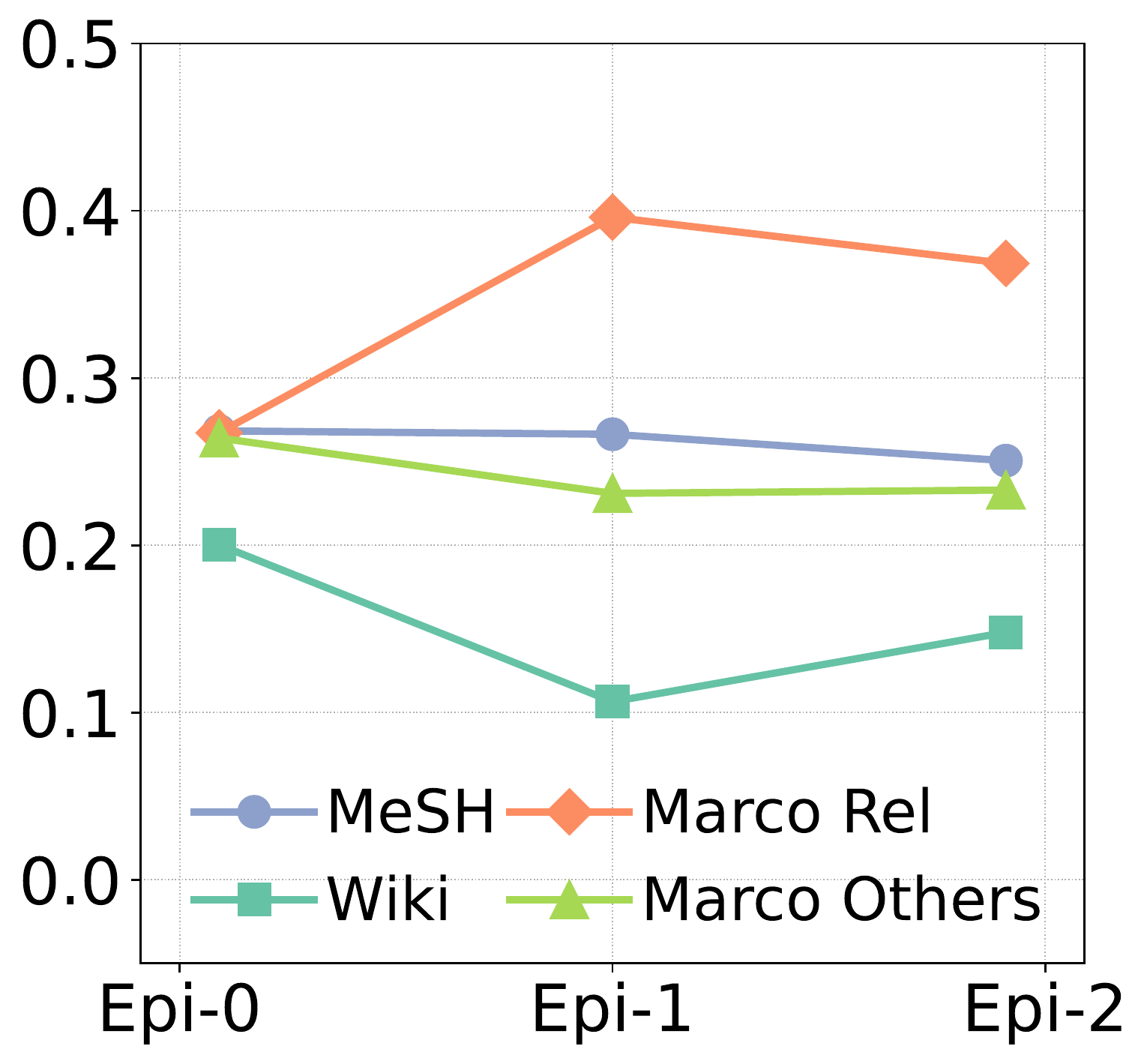}
     \caption{\model{} Att. Score. \label{subfig:moma_attn_score}}
\end{subfigure}
\begin{subfigure}[]{0.245\textwidth}
     \centering
     \includegraphics[width=\textwidth]{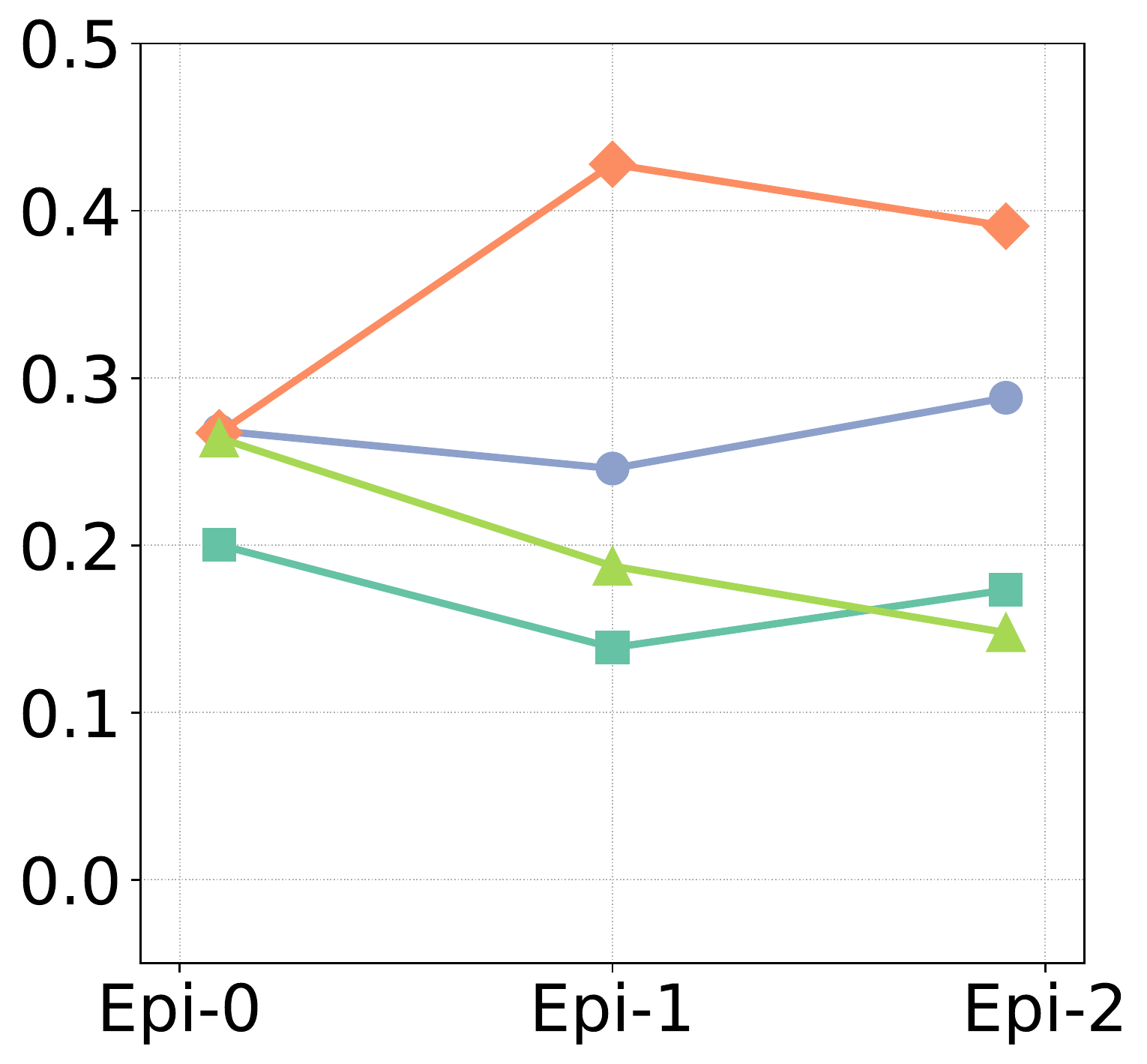}
     \caption{ADist Att. Score.}
\end{subfigure}
\begin{subfigure}[]{0.245\textwidth}
     \centering
     \includegraphics[width=\textwidth]{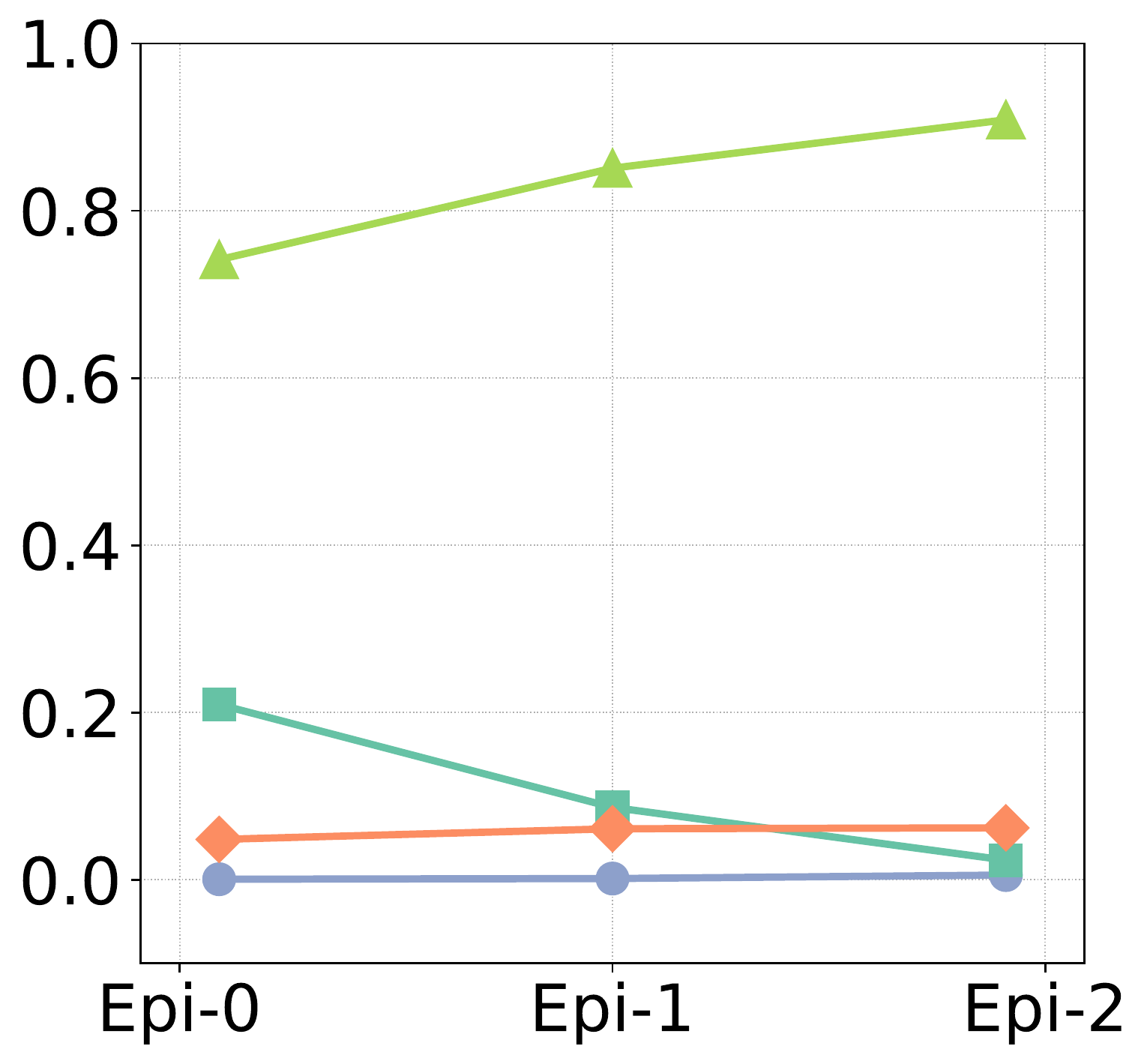}
     \caption{\model{} Doc Ratio.\label{subfig:moma_doc_ratio}}
\end{subfigure}
\begin{subfigure}[]{0.245\textwidth}
     \centering
     \includegraphics[width=\textwidth]{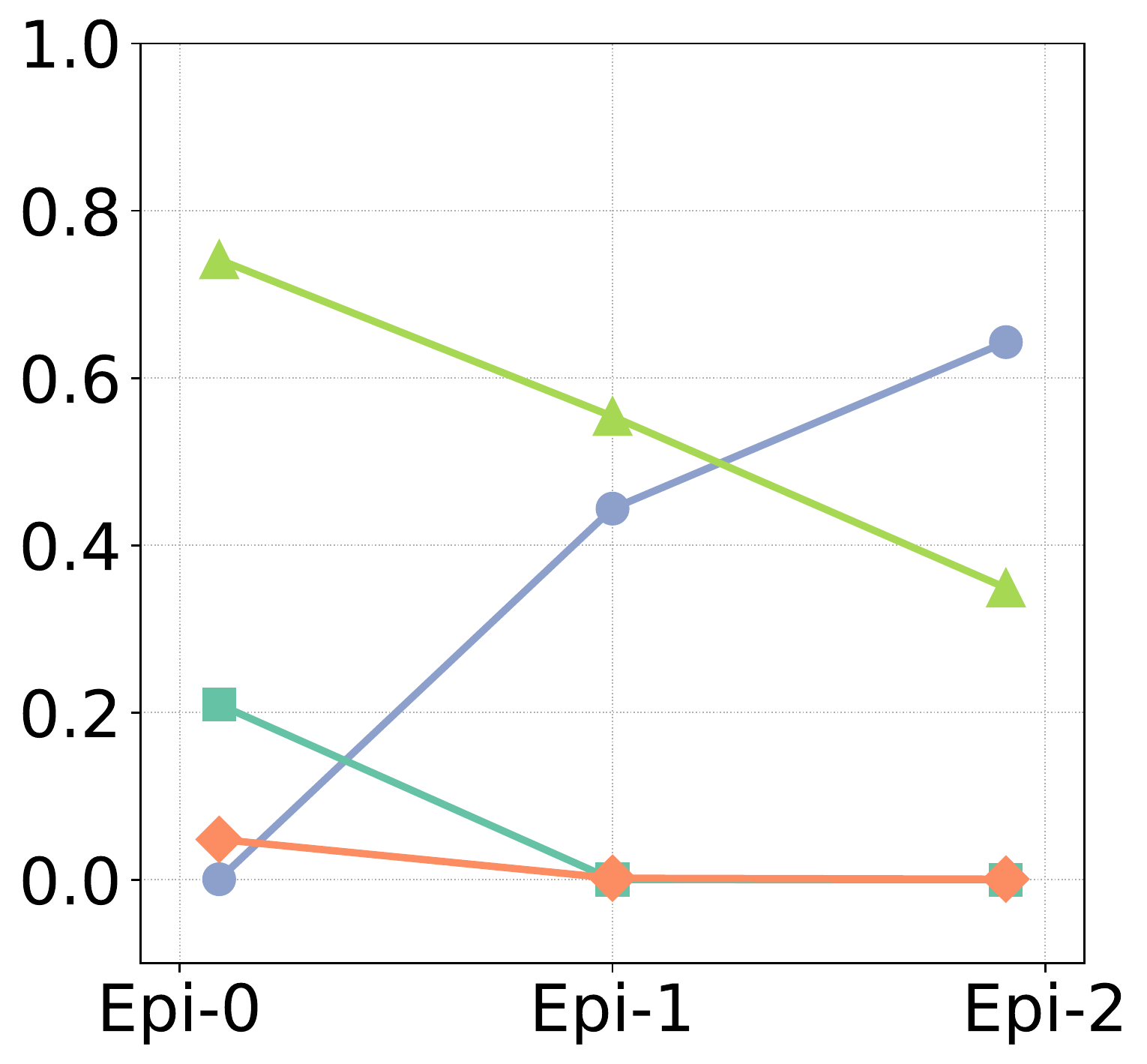}
     \caption{ADist Doc Ratio.}
\end{subfigure}
\caption{ Grounding component breakdown for different distillation methods in each learning iteration.
We display the regularized doc and att. score ratio of documents from different augmentation sources. \label{fig:learning}
}
\end{figure*}

\begin{figure*}[t]
\begin{subfigure}[]{0.245\textwidth}
     \centering
     \includegraphics[width=\textwidth]{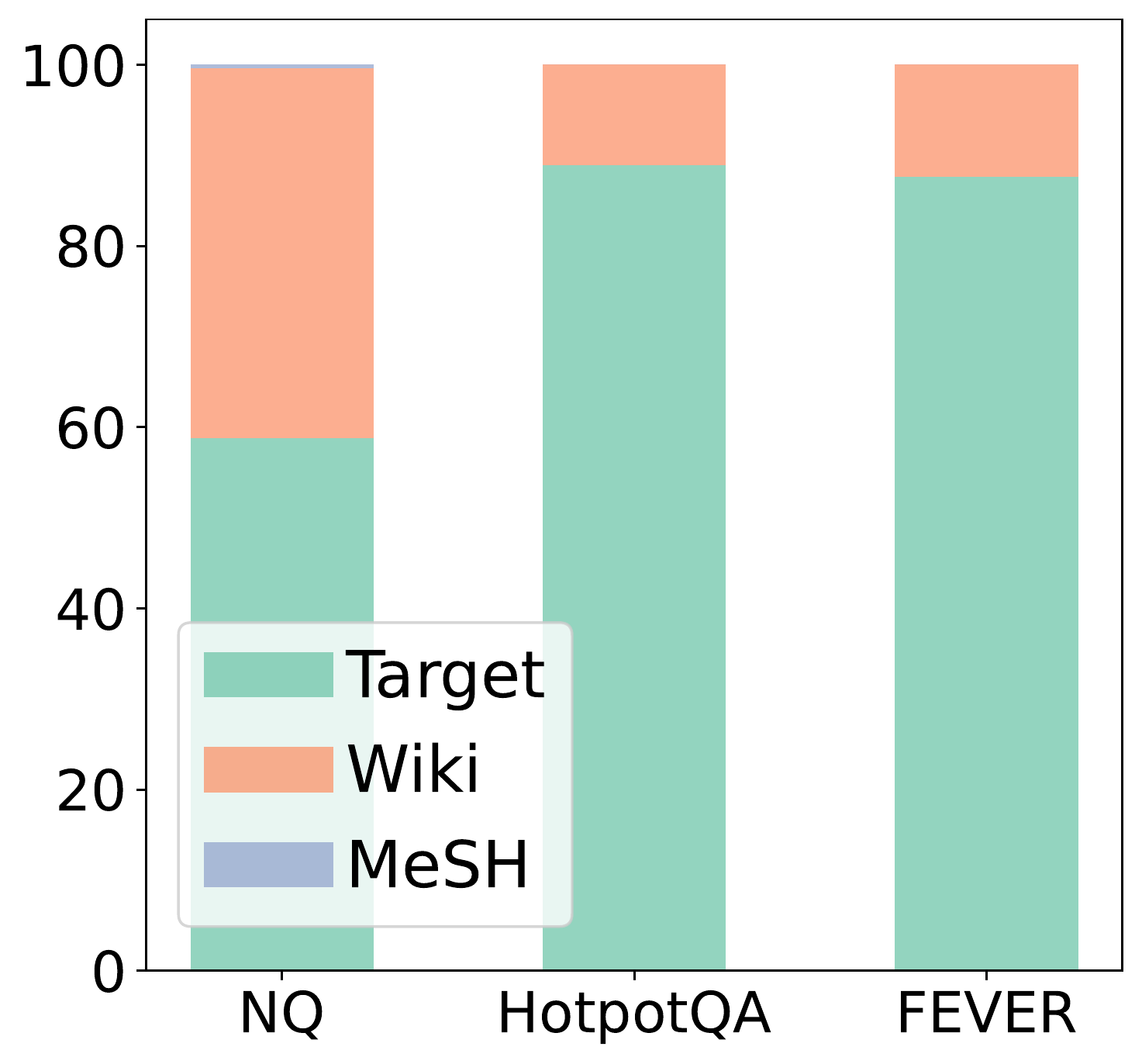}
     \caption{Doc Ratio. (Wiki)}
\end{subfigure}
\begin{subfigure}[]{0.245\textwidth}
     \centering
     \includegraphics[width=\textwidth]{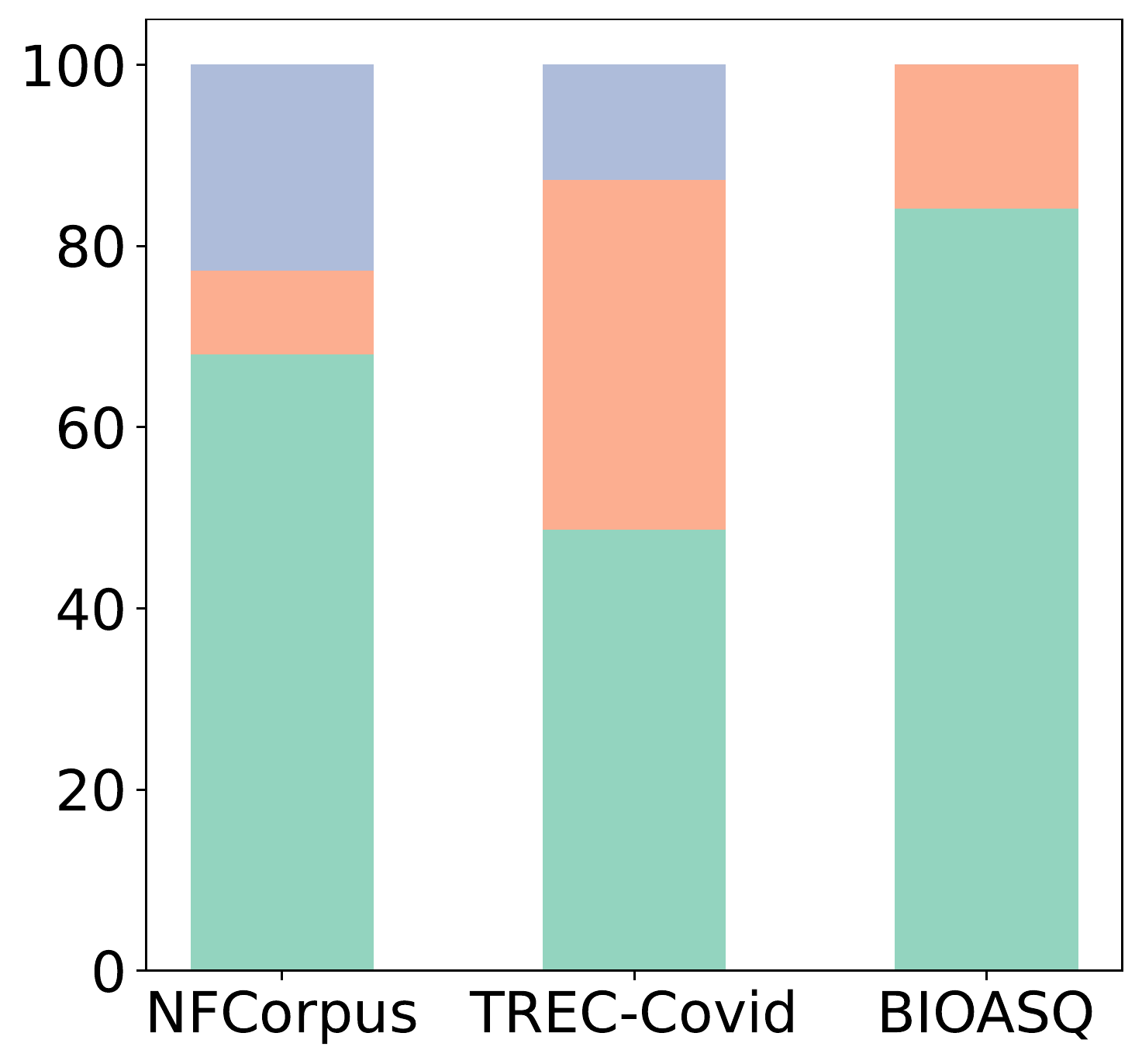}
     \caption{Doc Ratio. (Med)}
\end{subfigure}
\begin{subfigure}[]{0.245\textwidth}
     \centering
     \includegraphics[width=\textwidth]{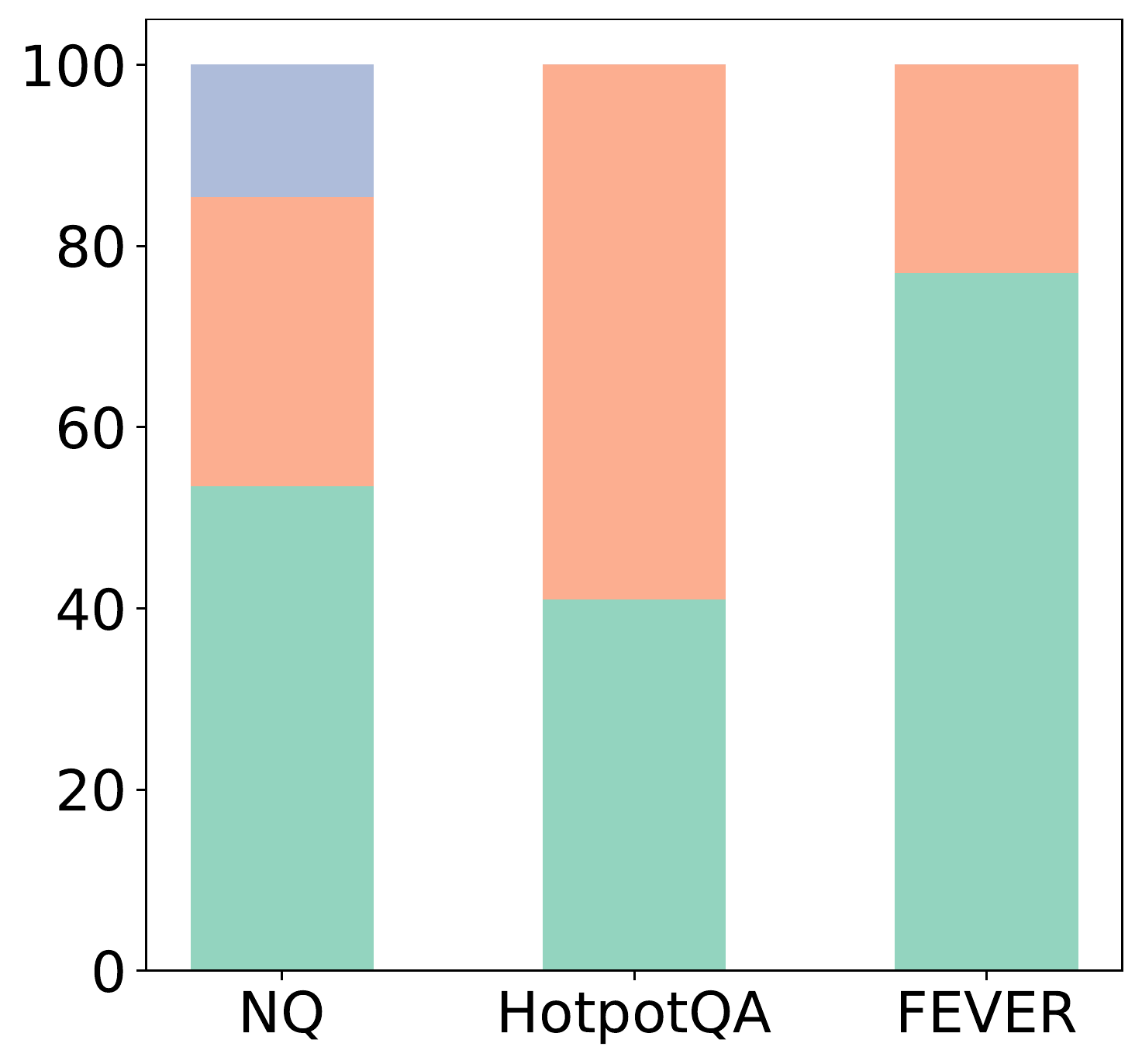}
     \caption{Att. Score Ratio. (Wiki)}
\end{subfigure}
\begin{subfigure}[]{0.245\textwidth}
     \centering
     \includegraphics[width=\textwidth]{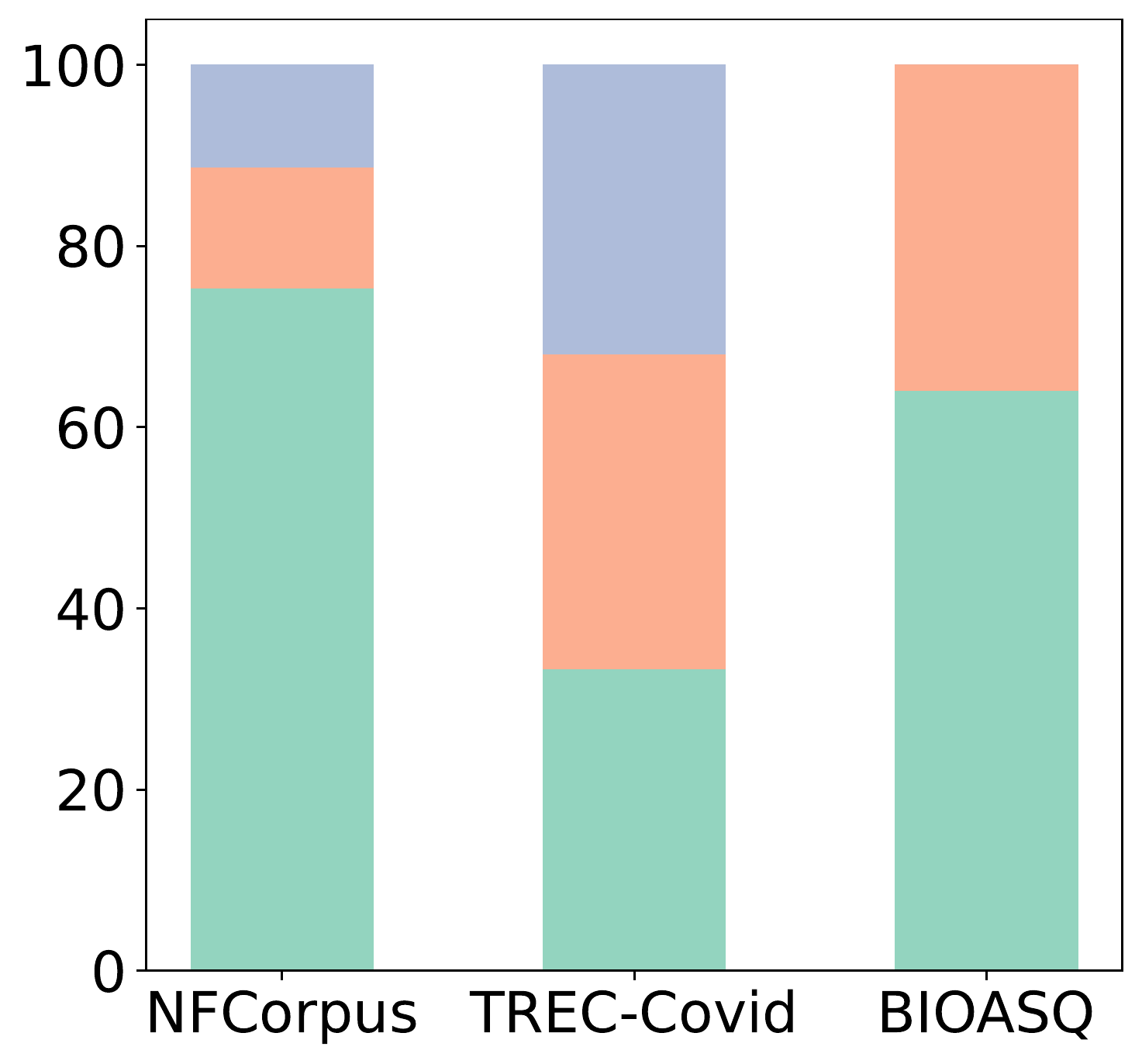}
     \caption{Att. Score Ratio. (Med)}
\end{subfigure}
\caption{The inclusion of Plug-In memory during testing (grouped by the Wiki and Medical domains). \label{fig:plugin}
}
\vspace{-0.3cm}
\end{figure*}


To study the effect of our joint learning mechanism on the memory mixture, we compare it with recent state-of-the-art Attention Distillation (ADist), which is first used in \citet{izacard2020distilling} and recently updated in a parallel work~\citet{izacard2022few}. 
It jointly trains the augmentation model using attention scores from the end language model as pseudo-labels.
We also enrich ADist with relevance labels from MARCO for more direct supervision, which was shown to be effective in distilling a dense retriever from stronger cross-encoder ranking model~\citep{hofstatter2021efficiently}.
Similar to previous section, to exclude the performance gain brought by contrastive pretraining, we choose \model{} (T5-ANCE) as our own method for comparison.
The performances of these joint learning methods are listed in Table~\ref{tab:distill}.
We pick six BEIR tasks whose domains are closely related to the augmentation corpora: TREC-COVID, BIOASQ, and NFCorpus are medical search and closely related to MeSH.
NQ, HotpotQA, and FEVER are all Wikipedia based.
The results show that ADist, either standalone or enriched with MARCO labels, does not improve the final accuracy compared to using a supervised dense retriever as the augmentation component without joint learning. The main difference is that the supervised retriever has been trained effectively using hard negative sampling~\citep{xiong2020approximate}.
Jointly learning using soft labels without hard negatives downgraded the augmentation accuracy. Hence, \model{} is a simple technique to learn the end task signals via the attention scores together with hard negatives, which improves quality over a supervised retriever alone.

To further illustrate the joint training process, we track the attention scores of documents from different memory sources as well as their ratio in the augmentation set in Figure~\ref{fig:learning}.
We also split MARCO documents by whether they are labeled as \textbf{Relevant (Rel)} for the corresponding query.


Firstly, \model{} learns to increasingly attend to, and retrieve, relevant documents from the memory mixture throughout training.
In Figure~\ref{subfig:moma_attn_score}, more attention is paid to MARCO Relevant documents than to any other type in the memory. Although the number of MARCO Relevant documents is not significant as a percentage of the augmenting set in Figure~\ref{subfig:moma_doc_ratio}, a query  level analysis confirms that percentage of queries having at least one relevant document in the augmenting set increases from 46\% in Epi-0 to 62\% in Epi-2. 

This apparent discrepancy can be explained by the fact that MARCO has only one relevant label per query on average, leaving plenty of room for other types of documents to be included in the augmenting set. 


Secondly, the amount of attention paid to certain types of documents by \model{} is positively correlated with their representation in the augmenting set. This confirms that the joint learning effectively conveys the feedback signals from the end model to the augmentation component. For instance, in Figure~\ref{subfig:moma_attn_score}, \model{} pays a high level of attention to MARCO Other documents, a signal reflected in the composition of its augmentation set in Figure~\ref{subfig:moma_doc_ratio}. Even though MARCO Other documents were not labeled relevant for the query, they can still prove to be valuable as an augmenting document because they may contain partial information that helps query understanding~\citep{lavrenko2017relevance} or it was simply not annotated in MARCO's sparse labels~\citep{msmarco}. In comparison, the correlation of the two in ADist is weak as the model seems to include 60\% augmenting documents from MeSH, far greater than the fraction of medical queries in MARCO.


\subsection{Generalization of Plug-In Memory}
\label{sec:exp_plugin}

In the previous section, we observed how \model{} learns to attend to, and retrieve, informative documents from memories on which it was trained. In this section, we examine the zero-shot behavior of \model{} (T5-ANCE) on new corpora plugged-in at test time (keeping Wiki and MeSH as before).

Figure~\ref{fig:plugin} compares documents from the plugged-in target versus the remaining memory mixture in terms of membership in the augmenting set (Doc Ratio) and attention. 
Again, on all tasks, \model{} (T5-ANCE) heavily attends to -- and successfully retrieves -- in-domain documents, even if those in-domain documents were only just plugged in. 
This confirms that the augmentation model achieves the zero-shot ability to capture relevant information from unseen corpora.

In the medical domain, the model pays more attention to MeSH documents, especially on TREC-Covid task since MeSH includes high quality updated information related to COVID-19.
Wikipedia documents received more attention on the Wiki-centric tasks like FEVER, as expected. Some tasks may need a small amount of precise information from Wikipedia to answer the detailed question, e.g. in HotpotQA.  
Similar with the training process, there is a non-trivial correspondence between attention score of a memory and its membership in the augmentation set.

\subsection{Case Studies}
\label{sec:case}
Table~\ref{tab:case} shows examples of how augmenting documents chosen by \model{} can provide valuable contextual information for the query. 
The first example is a training query from MARCO, where the augmenting documents help disambiguate the query word "rating". 
In the second one, documents from the official Wiki and HotpotQA's Wiki corpus are descriptions of the two entities in HotpotQA's comparison question. 
It illustrates how \model{} provides more comprehensive augmentation by incorporating information from different sources.

\begin{table}[t]
\centering
\small
\begin{tabularx}{\textwidth}{p{1.5cm}|X}
\toprule
\textbf{Queries} & \textbf{Augmentation Docs} \\ 
\hline
\multicolumn{2}{l}{\textbf{Training}} \\ \hline
\textbf{[Marco]} What is hotel transylvania rated
&\textbf{[Marco]} Why is Hotel Transylvania 2 rated PG? It is rated PG for some scary images, action and rude humor.
\textbf{[Wiki]} Another review aggregate calculated an average score of 47 out of 100, indicating ``mixed or average reviews''. \\
\hline
\multicolumn{2}{l}{\textbf{Zero-Shot Testing}} \\ \hline
\textbf{[HotpotQA]} Were Scott Derrickson and Ed Wood of the same nationality?
&\textbf{[Wiki]} Scott Derrickson (born July 16, 1966) is an American director, screenwriter and producer.
\textbf{[HotpotQA]}  Edward Davis Wood Jr. (October 10, December 10, 1978) was an American filmmaker, actor, writer, producer, and director.\\
\bottomrule
\end{tabularx}
\caption{\model{} retrieves augmenting documents during training (Marco) and testing (BEIR).
\label{tab:case}}
\vspace{-2ex}
\end{table}

\section{Conclusion}



In this paper we propose a new plug-in mixture-of-memory mechanism for the retrieval augmented language models to improve their zero-shot ability on the dense retrieval task.
To learn the memory mixture we develop a new joint learning approach that trains the augmentation component using the positive signals from the end task, the language model's attention scores, and hard negatives retrieved from the mixture of augmentation corpora.
This leads to our final model \model{} (T5-ANCE) and \model{} (COCO) that achieve strong zero-shot accuracy on 18 retrieval tasks included in BEIR.
Our analysis shows the importance of augmenting with diverse memory sources and  in-domain information for robust generalization.
We also share our observations and insights on how the model learns to leverage the augmentation information from multiple corpora during training and testing.
We hope our findings and illustrations can inspire more future research in better augmenting language models, to provide other alternatives to achieve generalization ability beyond solely relying on model scale.

\clearpage
\section*{Limitations}
Although \model (T5-ANCE) and \model (COCO) achieve strong zero-shot performances, we mainly verify their efficacy from the empirical performances on BEIR
tasks, where the target corpora, Wiki and MARCO serve as readily available retrieval sources.
In a real-world scenario, the grounding corpora usually need to be customized according to query domains and user needs.
Thus, how to choose effective grounding corpora and efficiently evaluate their relative contribution remain an open problem.
These analyses will go beyond our empirical settings and reveal a wider application scenario of {\model}.

\section*{Ethics Statement}
All data in this study are publicly available and used under ethical considerations. Text and figures in the paper are used for illustration only, they do not represent the ethical attitude of the authors.

\bibliography{iclr2023_conference}

\begin{thebibliography}{70}
\expandafter\ifx\csname natexlab\endcsname\relax\def\natexlab#1{#1}\fi

\bibitem[{Bajaj et~al.(2016)Bajaj, Campos, Craswell, Deng, Gao, Liu, Majumder,
  McNamara, Mitra, Nguyen et~al.}]{msmarco}
Payal Bajaj, Daniel Campos, Nick Craswell, Li~Deng, Jianfeng Gao, Xiaodong Liu,
  Rangan Majumder, Andrew McNamara, Bhaskar Mitra, Tri Nguyen, et~al. 2016.
\newblock {MS MARCO}: A human generated machine reading comprehension dataset.
\newblock \emph{arXiv preprint arXiv:1611.09268}.

\bibitem[{Bender et~al.(2021)Bender, Gebru, McMillan-Major, and
  Shmitchell}]{bender2021dangers}
Emily~M Bender, Timnit Gebru, Angelina McMillan-Major, and Shmargaret
  Shmitchell. 2021.
\newblock On the dangers of stochastic parrots: Can language models be too big?
\newblock In \emph{Proceedings of the 2021 ACM Conference on Fairness,
  Accountability, and Transparency}, pages 610--623.

\bibitem[{Bondarenko et~al.(2020)Bondarenko, Fr{\"o}be, Beloucif, Gienapp,
  Ajjour, Panchenko, Biemann, Stein, Wachsmuth, Potthast, and Hagen}]{touche}
Alexander Bondarenko, Maik Fr{\"o}be, Meriem Beloucif, Lukas Gienapp, Yamen
  Ajjour, Alexander Panchenko, Chris Biemann, Benno Stein, Henning Wachsmuth,
  Martin Potthast, and Matthias Hagen. 2020.
\newblock \href {http://ceur-ws.org/Vol-2696/} {{Overview of Touch{\'e} 2020:
  Argument Retrieval}}.
\newblock In \emph{Working Notes Papers of the CLEF 2020 Evaluation Labs},
  volume 2696 of \emph{CEUR Workshop Proceedings}.

\bibitem[{Borgeaud et~al.(2022)Borgeaud, Mensch, Hoffmann, Cai, Rutherford,
  Millican, Van Den~Driessche, Lespiau, Damoc, Clark
  et~al.}]{borgeaud2022improving}
Sebastian Borgeaud, Arthur Mensch, Jordan Hoffmann, Trevor Cai, Eliza
  Rutherford, Katie Millican, George~Bm Van Den~Driessche, Jean-Baptiste
  Lespiau, Bogdan Damoc, Aidan Clark, et~al. 2022.
\newblock Improving language models by retrieving from trillions of tokens.
\newblock In \emph{International Conference on Machine Learning}, pages
  2206--2240. PMLR.

\bibitem[{Boteva et~al.(2016)Boteva, Gholipour, Sokolov, and
  Riezler}]{nfcorpus}
Vera Boteva, Demian Gholipour, Artem Sokolov, and Stefan Riezler. 2016.
\newblock A full-text learning to rank dataset for medical information
  retrieval.
\newblock In \emph{European Conference on Information Retrieval}, pages
  716--722. Springer.

\bibitem[{Brown et~al.(2020)Brown, Mann, Ryder, Subbiah, Kaplan, Dhariwal,
  Neelakantan, Shyam, Sastry, Askell et~al.}]{brown2020language}
Tom Brown, Benjamin Mann, Nick Ryder, Melanie Subbiah, Jared~D Kaplan, Prafulla
  Dhariwal, Arvind Neelakantan, Pranav Shyam, Girish Sastry, Amanda Askell,
  et~al. 2020.
\newblock Language models are few-shot learners.
\newblock \emph{Advances in neural information processing systems},
  33:1877--1901.

\bibitem[{Carpineto and Romano(2012)}]{carpineto2012survey}
Claudio Carpineto and Giovanni Romano. 2012.
\newblock A survey of automatic query expansion in information retrieval.
\newblock \emph{Acm Computing Surveys (CSUR)}, 44(1):1--50.

\bibitem[{Chen et~al.(2017)Chen, Fisch, Weston, and Bordes}]{chen2017reading}
Danqi Chen, Adam Fisch, Jason Weston, and Antoine Bordes. 2017.
\newblock Reading {W}ikipedia to {A}nswer {O}pen-{D}omain {Q}uestions.
\newblock In \emph{Proceedings of the 55th Annual Meeting of the Association
  for Computational Linguistics}, pages 1870--1879.

\bibitem[{Cohan et~al.(2020)Cohan, Feldman, Beltagy, Downey, and
  Weld}]{scidocs}
Arman Cohan, Sergey Feldman, Iz~Beltagy, Doug Downey, and Daniel Weld. 2020.
\newblock \href {https://aclanthology.org/2020.acl-main.207} {{SPECTER}:
  Document-level representation learning using citation-informed transformers}.
\newblock In \emph{Proceedings of the 58th Annual Meeting of the Association
  for Computational Linguistics}, pages 2270--2282, Online. Association for
  Computational Linguistics.

\bibitem[{Diggelmann et~al.(2020)Diggelmann, Boyd-Graber, Bulian, Ciaramita,
  and Leippold}]{climatefever}
Thomas Diggelmann, Jordan Boyd-Graber, Jannis Bulian, Massimiliano Ciaramita,
  and Markus Leippold. 2020.
\newblock {CLIMATE-FEVER}: A dataset for verification of real-world climate
  claims.
\newblock \emph{arXiv preprint arXiv:2012.00614}.

\bibitem[{Formal et~al.(2021)Formal, Piwowarski, and
  Clinchant}]{formal2021splade}
Thibault Formal, Benjamin Piwowarski, and St{\'e}phane Clinchant. 2021.
\newblock Splade: Sparse lexical and expansion model for first stage ranking.
\newblock In \emph{Proceedings of the 44th International ACM SIGIR Conference
  on Research and Development in Information Retrieval}, pages 2288--2292.

\bibitem[{Gao and Callan(2022)}]{gao2022cocondenser}
Luyu Gao and Jamie Callan. 2022.
\newblock Unsupervised corpus aware language model pre-training for dense
  passage retrieval.
\newblock In \emph{ACL 2022}.

\bibitem[{Guu et~al.(2020)Guu, Lee, Tung, Pasupat, and Chang}]{guu2020realm}
Kelvin Guu, Kenton Lee, Zora Tung, Panupong Pasupat, and Ming-Wei Chang. 2020.
\newblock {REALM}: Retrieval-augmented language model pre-training.
\newblock In \emph{ICML}.

\bibitem[{Hasibi et~al.(2017)Hasibi, Nikolaev, Xiong, Balog, Bratsberg, Kotov,
  and Callan}]{dbpedia}
Faegheh Hasibi, Fedor Nikolaev, Chenyan Xiong, Krisztian Balog, Svein~Erik
  Bratsberg, Alexander Kotov, and Jamie Callan. 2017.
\newblock \href {https://doi.org/10.1145/3077136.3080751} {{DBpedia-Entity v2}:
  A test collection for entity search}.
\newblock In \emph{Proceedings of the 40th International ACM SIGIR Conference
  on Research and Development in Information Retrieval}, SIGIR '17, page
  1265–1268, New York, NY, USA. Association for Computing Machinery.

\bibitem[{Hoffmann et~al.(2022)Hoffmann, Borgeaud, Mensch, Buchatskaya, Cai,
  Rutherford, Casas, Hendricks, Welbl, Clark et~al.}]{hoffmann2022training}
Jordan Hoffmann, Sebastian Borgeaud, Arthur Mensch, Elena Buchatskaya, Trevor
  Cai, Eliza Rutherford, Diego de~Las Casas, Lisa~Anne Hendricks, Johannes
  Welbl, Aidan Clark, et~al. 2022.
\newblock Training compute-optimal large language models.
\newblock \emph{arXiv preprint arXiv:2203.15556}.

\bibitem[{Hofst{\"a}tter et~al.(2021)Hofst{\"a}tter, Lin, Yang, Lin, and
  Hanbury}]{hofstatter2021efficiently}
Sebastian Hofst{\"a}tter, Sheng-Chieh Lin, Jheng-Hong Yang, Jimmy Lin, and
  Allan Hanbury. 2021.
\newblock \href
  {https://dl.acm.org/doi/abs/10.1145/3404835.3462891?casa_token=E1-QtAoihwgAAAAA:LXH6aLlYff3ScPyFzv560IyfHwR_EAMOHRY_rDY9dzL8q9qi7Dm8Z_M1YyUXdc6pMjomLQI5OlRiO9A}
  {Efficiently teaching an effective dense retriever with balanced topic aware
  sampling}.
\newblock In \emph{Proceedings of SIGIR 2021}, pages 113--122.

\bibitem[{Hofst\"{a}tter et~al.(2021)Hofst\"{a}tter, Lin, Yang, Lin, and
  Hanbury}]{tasb}
Sebastian Hofst\"{a}tter, Sheng-Chieh Lin, Jheng-Hong Yang, Jimmy Lin, and
  Allan Hanbury. 2021.
\newblock \href {https://doi.org/10.1145/3404835.3462891} {Efficiently teaching
  an effective dense retriever with balanced topic aware sampling}.
\newblock In \emph{Proceedings of the 44th International ACM SIGIR Conference
  on Research and Development in Information Retrieval}, page 113–122.
  Association for Computing Machinery.

\bibitem[{Hoogeveen et~al.(2015)Hoogeveen, Verspoor, and Baldwin}]{cqadupstack}
Doris Hoogeveen, Karin~M. Verspoor, and Timothy Baldwin. 2015.
\newblock \href {https://doi.org/10.1145/2838931.2838934} {{CQADupStack}: A
  benchmark data set for community question-answering research}.
\newblock In \emph{Proceedings of the 20th Australasian Document Computing
  Symposium}, ADCS '15, New York, NY, USA. Association for Computing Machinery.

\bibitem[{Izacard et~al.(2021)Izacard, Caron, Hosseini, Riedel, Bojanowski,
  Joulin, and Grave}]{izacard2021towards}
Gautier Izacard, Mathilde Caron, Lucas Hosseini, Sebastian Riedel, Piotr
  Bojanowski, Armand Joulin, and Edouard Grave. 2021.
\newblock Towards unsupervised dense information retrieval with contrastive
  learning.
\newblock \emph{arXiv preprint arXiv:2112.09118}.

\bibitem[{Izacard and Grave(2020{\natexlab{a}})}]{izacard2020distilling}
Gautier Izacard and Edouard Grave. 2020{\natexlab{a}}.
\newblock \href {https://arxiv.org/abs/2012.04584} {Distilling knowledge from
  reader to retriever for question answering}.
\newblock \emph{arXiv preprint arXiv:2012.04584}.

\bibitem[{Izacard and Grave(2020{\natexlab{b}})}]{izacard2020leveraging}
Gautier Izacard and Edouard Grave. 2020{\natexlab{b}}.
\newblock \href {https://arxiv.org/abs/2007.0128} {Leveraging passage retrieval
  with generative models for open domain question answering}.

\bibitem[{Izacard et~al.(2022)Izacard, Lewis, Lomeli, Hosseini, Petroni,
  Schick, Dwivedi-Yu, Joulin, Riedel, and Grave}]{izacard2022few}
Gautier Izacard, Patrick Lewis, Maria Lomeli, Lucas Hosseini, Fabio Petroni,
  Timo Schick, Jane Dwivedi-Yu, Armand Joulin, Sebastian Riedel, and Edouard
  Grave. 2022.
\newblock Few-shot learning with retrieval augmented language models.
\newblock \emph{arXiv preprint arXiv:2208.03299}.

\bibitem[{Johnson et~al.(2019)Johnson, Douze, and
  J{\'e}gou}]{johnson2019billion}
Jeff Johnson, Matthijs Douze, and Herv{\'e} J{\'e}gou. 2019.
\newblock Billion-scale similarity search with gpus.
\newblock \emph{IEEE Transactions on Big Data}, 7(3):535--547.

\bibitem[{Kaplan et~al.(2020)Kaplan, McCandlish, Henighan, Brown, Chess, Child,
  Gray, Radford, Wu, and Amodei}]{kaplan2020scaling}
Jared Kaplan, Sam McCandlish, Tom Henighan, Tom~B Brown, Benjamin Chess, Rewon
  Child, Scott Gray, Alec Radford, Jeffrey Wu, and Dario Amodei. 2020.
\newblock Scaling laws for neural language models.
\newblock \emph{arXiv preprint arXiv:2001.08361}.

\bibitem[{Karpukhin et~al.(2020)Karpukhin, O{\u{g}}uz, Min, Lewis, Wu, Edunov,
  Chen, and Yih}]{karpukhin2020dense}
Vladimir Karpukhin, Barlas O{\u{g}}uz, Sewon Min, Patrick Lewis, Ledell Wu,
  Sergey Edunov, Danqi Chen, and Wen-tau Yih. 2020.
\newblock Dense passage retrieval for open-domain question answering.
\newblock \emph{arXiv preprint arXiv:2004.04906}.

\bibitem[{Khandelwal et~al.(2019)Khandelwal, Levy, Jurafsky, Zettlemoyer, and
  Lewis}]{khandelwal2019generalization}
Urvashi Khandelwal, Omer Levy, Dan Jurafsky, Luke Zettlemoyer, and Mike Lewis.
  2019.
\newblock Generalization through memorization: Nearest neighbor language
  models.
\newblock \emph{arXiv preprint arXiv:1911.00172}.

\bibitem[{Khattab and Zaharia(2020{\natexlab{a}})}]{khattab2020colbert}
Omar Khattab and Matei Zaharia. 2020{\natexlab{a}}.
\newblock Colbert: Efficient and effective passage search via contextualized
  late interaction over bert.
\newblock In \emph{Proceedings of the 43rd International ACM SIGIR conference
  on research and development in Information Retrieval}, pages 39--48.

\bibitem[{Khattab and Zaharia(2020{\natexlab{b}})}]{colbert}
Omar Khattab and Matei Zaharia. 2020{\natexlab{b}}.
\newblock \href {https://doi.org/10.1145/3397271.3401075} {Colbert: Efficient
  and effective passage search via contextualized late interaction over bert}.
\newblock In \emph{Proceedings of the 43rd International ACM SIGIR Conference
  on Research and Development in Information Retrieval}, page 39–48, New
  York, NY, USA. Association for Computing Machinery.

\bibitem[{Kim(2022)}]{kim2022applications}
Yubin Kim. 2022.
\newblock Applications and future of dense retrieval in industry.
\newblock In \emph{Proceedings of the 45th International ACM SIGIR Conference
  on Research and Development in Information Retrieval}, pages 3373--3374.

\bibitem[{Kwiatkowski et~al.(2019)Kwiatkowski, Palomaki, Redfield, Collins,
  Parikh, Alberti, Epstein, Polosukhin, Devlin, Lee, Toutanova, Jones, Kelcey,
  Chang, Dai, Uszkoreit, Le, and Petrov}]{nq}
Tom Kwiatkowski, Jennimaria Palomaki, Olivia Redfield, Michael Collins, Ankur
  Parikh, Chris Alberti, Danielle Epstein, Illia Polosukhin, Jacob Devlin,
  Kenton Lee, Kristina Toutanova, Llion Jones, Matthew Kelcey, Ming-Wei Chang,
  Andrew~M. Dai, Jakob Uszkoreit, Quoc Le, and Slav Petrov. 2019.
\newblock \href {https://aclanthology.org/Q19-1026} {Natural questions: A
  benchmark for question answering research}.
\newblock \emph{Transactions of the Association for Computational Linguistics},
  7:452--466.

\bibitem[{Lavrenko and Croft(2017)}]{lavrenko2017relevance}
Victor Lavrenko and W~Bruce Croft. 2017.
\newblock Relevance-based language models.
\newblock In \emph{ACM SIGIR Forum}, volume~51, pages 260--267. ACM New York,
  NY, USA.

\bibitem[{Lewis et~al.(2020)Lewis, Perez, Piktus, Petroni, Karpukhin, Goyal,
  K{\"u}ttler, Lewis, Yih, Rockt{\"a}schel et~al.}]{lewis2020retrieval}
Patrick Lewis, Ethan Perez, Aleksandara Piktus, Fabio Petroni, Vladimir
  Karpukhin, Naman Goyal, Heinrich K{\"u}ttler, Mike Lewis, Wen-tau Yih, Tim
  Rockt{\"a}schel, et~al. 2020.
\newblock Retrieval-augmented generation for knowledge-intensive nlp tasks.
\newblock \emph{arXiv preprint arXiv:2005.11401}.

\bibitem[{Liu et~al.(2019)Liu, Ott, Goyal, Du, Joshi, Chen, Levy, Lewis,
  Zettlemoyer, and Stoyanov}]{liu2019roberta}
Yinhan Liu, Myle Ott, Naman Goyal, Jingfei Du, Mandar Joshi, Danqi Chen, Omer
  Levy, Mike Lewis, Luke Zettlemoyer, and Veselin Stoyanov. 2019.
\newblock Ro{BERT}a: {A} {R}obustly {O}ptimized {BERT} {P}retraining
  {A}pproach.
\newblock \emph{arXiv preprint arXiv:1907.11692}.

\bibitem[{Loshchilov and Hutter(2019)}]{loshchilov2018adamw}
Ilya Loshchilov and Frank Hutter. 2019.
\newblock \href {https://openreview.net/forum?id=Bkg6RiCqY7} {Decoupled weight
  decay regularization}.
\newblock In \emph{International Conference on Learning Representations}.

\bibitem[{Lu et~al.(2021)Lu, Hernandez~Abrego, Ma, Ni, and
  Yang}]{lu-etal-2021-multi}
Jing Lu, Gustavo Hernandez~Abrego, Ji~Ma, Jianmo Ni, and Yinfei Yang. 2021.
\newblock \href {https://doi.org/10.18653/v1/2021.emnlp-main.492} {Multi-stage
  training with improved negative contrast for neural passage retrieval}.
\newblock In \emph{Proceedings of the 2021 Conference on Empirical Methods in
  Natural Language Processing}, pages 6091--6103, Online and Punta Cana,
  Dominican Republic. Association for Computational Linguistics.

\bibitem[{Maia et~al.(2018)Maia, Handschuh, Freitas, Davis, McDermott, Zarrouk,
  and Balahur}]{fiqa}
Macedo Maia, Siegfried Handschuh, Andr\'{e} Freitas, Brian Davis, Ross
  McDermott, Manel Zarrouk, and Alexandra Balahur. 2018.
\newblock \href {https://doi.org/10.1145/3184558.3192301} {{WWW}'18 open
  challenge: Financial opinion mining and question answering}.
\newblock In \emph{Companion Proceedings of the The Web Conference 2018}, WWW
  '18, page 1941–1942, Republic and Canton of Geneva, CHE. International
  World Wide Web Conferences Steering Committee.

\bibitem[{Neelakantan et~al.(2022)Neelakantan, Xu, Puri, Radford, Han, Tworek,
  Yuan, Tezak, Kim, Hallacy et~al.}]{cpt}
Arvind Neelakantan, Tao Xu, Raul Puri, Alec Radford, Jesse~Michael Han, Jerry
  Tworek, Qiming Yuan, Nikolas Tezak, Jong~Wook Kim, Chris Hallacy, et~al.
  2022.
\newblock Text and code embeddings by contrastive pre-training.
\newblock \emph{arXiv preprint arXiv:2201.10005}.

\bibitem[{Ni et~al.(2022)Ni, Abrego, Constant, Ma, Hall, Cer, and
  Yang}]{ni2022sentence}
Jianmo Ni, Gustavo~Hernandez Abrego, Noah Constant, Ji~Ma, Keith Hall, Daniel
  Cer, and Yinfei Yang. 2022.
\newblock Sentence-t5: Scalable sentence encoders from pre-trained text-to-text
  models.
\newblock In \emph{Findings of the Association for Computational Linguistics:
  ACL 2022}, pages 1864--1874.

\bibitem[{Ni et~al.(2021)Ni, Qu, Lu, Dai, {\'A}brego, Ma, Zhao, Luan, Hall,
  Chang et~al.}]{gtr}
Jianmo Ni, Chen Qu, Jing Lu, Zhuyun Dai, Gustavo~Hern{\'a}ndez {\'A}brego,
  Ji~Ma, Vincent~Y Zhao, Yi~Luan, Keith~B Hall, Ming-Wei Chang, et~al. 2021.
\newblock Large dual encoders are generalizable retrievers.
\newblock \emph{arXiv preprint arXiv:2112.07899}.

\bibitem[{Paszke et~al.(2019)Paszke, Gross, Massa, Lerer, Bradbury, Chanan,
  Killeen, Lin, Gimelshein, Antiga et~al.}]{paszke2019pytorch}
Adam Paszke, Sam Gross, Francisco Massa, Adam Lerer, James Bradbury, Gregory
  Chanan, Trevor Killeen, Zeming Lin, Natalia Gimelshein, Luca Antiga, et~al.
  2019.
\newblock Pytorch: An imperative style, high-performance deep learning library.
\newblock \emph{Advances in neural information processing systems}, 32.

\bibitem[{Petroni et~al.(2020)Petroni, Piktus, Fan, Lewis, Yazdani, De~Cao,
  Thorne, Jernite, Karpukhin, Maillard et~al.}]{petroni2020kilt}
Fabio Petroni, Aleksandra Piktus, Angela Fan, Patrick Lewis, Majid Yazdani,
  Nicola De~Cao, James Thorne, Yacine Jernite, Vladimir Karpukhin, Jean
  Maillard, et~al. 2020.
\newblock Kilt: a benchmark for knowledge intensive language tasks.
\newblock \emph{arXiv preprint arXiv:2009.02252}.

\bibitem[{Qu et~al.(2021)Qu, Ding, Liu, Liu, Ren, Zhao, Dong, Wu, and
  Wang}]{rocketqa}
Yingqi Qu, Yuchen Ding, Jing Liu, Kai Liu, Ruiyang Ren, Wayne~Xin Zhao, Daxiang
  Dong, Hua Wu, and Haifeng Wang. 2021.
\newblock \href {https://aclanthology.org/2021.naacl-main.466} {{R}ocket{QA}:
  An optimized training approach to dense passage retrieval for open-domain
  question answering}.
\newblock In \emph{Proceedings of the 2021 Conference of the North American
  Chapter of the Association for Computational Linguistics: Human Language
  Technologies}, pages 5835--5847, Online. Association for Computational
  Linguistics.

\bibitem[{Raffel et~al.(2019)Raffel, Shazeer, Roberts, Lee, Narang, Matena,
  Zhou, Li, and Liu}]{raffel2019t5}
Colin Raffel, Noam Shazeer, Adam Roberts, Katherine Lee, Sharan Narang, Michael
  Matena, Yanqi Zhou, Wei Li, and Peter~J Liu. 2019.
\newblock Exploring the limits of transfer learning with a unified text-to-text
  transformer.
\newblock \emph{Journal of Machine Learning Research}.

\bibitem[{Roberts et~al.(2020)Roberts, Raffel, and Shazeer}]{roberts2020much}
Adam Roberts, Colin Raffel, and Noam Shazeer. 2020.
\newblock How much knowledge can you pack into the parameters of a language
  model?
\newblock In \emph{EMNLP}.

\bibitem[{Robertson et~al.(2009)Robertson, Zaragoza et~al.}]{bm25}
Stephen Robertson, Hugo Zaragoza, et~al. 2009.
\newblock The probabilistic relevance framework: Bm25 and beyond.
\newblock \emph{Foundations and Trends in Information Retrieval},
  3(4):333--389.

\bibitem[{Smith et~al.(2022)Smith, Patwary, Norick, LeGresley, Rajbhandari,
  Casper, Liu, Prabhumoye, Zerveas, Korthikanti et~al.}]{smith2022using}
Shaden Smith, Mostofa Patwary, Brandon Norick, Patrick LeGresley, Samyam
  Rajbhandari, Jared Casper, Zhun Liu, Shrimai Prabhumoye, George Zerveas,
  Vijay Korthikanti, et~al. 2022.
\newblock Using deepspeed and megatron to train megatron-turing nlg 530b, a
  large-scale generative language model.
\newblock \emph{arXiv preprint arXiv:2201.11990}.

\bibitem[{Soboroff et~al.(2018)Soboroff, Huang, and Harman}]{trecnews}
Ian Soboroff, Shudong Huang, and Donna Harman. 2018.
\newblock Trec 2018 news track overview.

\bibitem[{Strubell et~al.(2020)Strubell, Ganesh, and
  McCallum}]{strubell2020energy}
Emma Strubell, Ananya Ganesh, and Andrew McCallum. 2020.
\newblock Energy and policy considerations for modern deep learning research.
\newblock In \emph{Proceedings of the AAAI Conference on Artificial
  Intelligence}, volume~34, pages 13693--13696.

\bibitem[{Suarez et~al.(2018)Suarez, Albakour, Corney, Martinez, and
  Esquivel}]{signal1m}
Axel Suarez, Dyaa Albakour, David Corney, Miguel Martinez, and Jos{\'e}
  Esquivel. 2018.
\newblock A data collection for evaluating the retrieval of related tweets to
  news articles.
\newblock In \emph{European Conference on Information Retrieval}, pages
  780--786. Springer.

\bibitem[{Thakur et~al.(2021{\natexlab{a}})Thakur, Reimers, R{\"u}ckl{\'e},
  Srivastava, and Gurevych}]{thakur2021beir}
Nandan Thakur, Nils Reimers, Andreas R{\"u}ckl{\'e}, Abhishek Srivastava, and
  Iryna Gurevych. 2021{\natexlab{a}}.
\newblock Beir: A heterogenous benchmark for zero-shot evaluation of
  information retrieval models.
\newblock \emph{arXiv preprint arXiv:2104.08663}.

\bibitem[{Thakur et~al.(2021{\natexlab{b}})Thakur, Reimers, R{\"u}ckl{\'e},
  Srivastava, and Gurevych}]{beir}
Nandan Thakur, Nils Reimers, Andreas R{\"u}ckl{\'e}, Abhishek Srivastava, and
  Iryna Gurevych. 2021{\natexlab{b}}.
\newblock {BEIR}: A heterogenous benchmark for zero-shot evaluation of
  information retrieval models.
\newblock \emph{arXiv preprint arXiv:2104.08663}.

\bibitem[{Thorne et~al.(2018)Thorne, Vlachos, Christodoulopoulos, and
  Mittal}]{fever}
James Thorne, Andreas Vlachos, Christos Christodoulopoulos, and Arpit Mittal.
  2018.
\newblock \href {https://aclanthology.org/N18-1074} {{FEVER}: a large-scale
  dataset for fact extraction and {VER}ification}.
\newblock In \emph{Proceedings of the 2018 Conference of the North {A}merican
  Chapter of the Association for Computational Linguistics: Human Language
  Technologies, Volume 1 (Long Papers)}, pages 809--819, New Orleans,
  Louisiana. Association for Computational Linguistics.

\bibitem[{Tsatsaronis et~al.(2015)Tsatsaronis, Balikas, Malakasiotis, Partalas,
  Zschunke, Alvers, Weissenborn, Krithara, Petridis, Polychronopoulos
  et~al.}]{bioasq}
George Tsatsaronis, Georgios Balikas, Prodromos Malakasiotis, Ioannis Partalas,
  Matthias Zschunke, Michael~R Alvers, Dirk Weissenborn, Anastasia Krithara,
  Sergios Petridis, Dimitris Polychronopoulos, et~al. 2015.
\newblock An overview of the {BIOASQ} large-scale biomedical semantic indexing
  and question answering competition.
\newblock \emph{BMC bioinformatics}, 16(1):1--28.

\bibitem[{Voorhees et~al.(2021)Voorhees, Alam, Bedrick, Demner-Fushman, Hersh,
  Lo, Roberts, Soboroff, and Wang}]{treccovid}
Ellen Voorhees, Tasmeer Alam, Steven Bedrick, Dina Demner-Fushman, William~R.
  Hersh, Kyle Lo, Kirk Roberts, Ian Soboroff, and Lucy~Lu Wang. 2021.
\newblock \href {https://doi.org/10.1145/3451964.3451965} {{TREC-COVID}:
  Constructing a pandemic information retrieval test collection}.
\newblock \emph{SIGIR Forum}, 54(1).

\bibitem[{Voorhees et~al.(2004)}]{robust04}
Ellen~M Voorhees et~al. 2004.
\newblock Overview of the trec 2004 robust retrieval track.
\newblock In \emph{Trec}, pages 69--77.

\bibitem[{Wachsmuth et~al.(2018)Wachsmuth, Syed, and Stein}]{arguana}
Henning Wachsmuth, Shahbaz Syed, and Benno Stein. 2018.
\newblock \href {https://aclanthology.org/P18-1023} {Retrieval of the best
  counterargument without prior topic knowledge}.
\newblock In \emph{Proceedings of the 56th Annual Meeting of the Association
  for Computational Linguistics (Volume 1: Long Papers)}, pages 241--251,
  Melbourne, Australia. Association for Computational Linguistics.

\bibitem[{Wadden et~al.(2020)Wadden, Lin, Lo, Wang, van Zuylen, Cohan, and
  Hajishirzi}]{scifact}
David Wadden, Shanchuan Lin, Kyle Lo, Lucy~Lu Wang, Madeleine van Zuylen, Arman
  Cohan, and Hannaneh Hajishirzi. 2020.
\newblock \href {https://aclanthology.org/2020.emnlp-main.609} {Fact or
  fiction: Verifying scientific claims}.
\newblock In \emph{Proceedings of the 2020 Conference on Empirical Methods in
  Natural Language Processing (EMNLP)}, pages 7534--7550, Online. Association
  for Computational Linguistics.

\bibitem[{Wang et~al.(2022)Wang, Thakur, Reimers, and Gurevych}]{wang2021gpl}
Kexin Wang, Nandan Thakur, Nils Reimers, and Iryna Gurevych. 2022.
\newblock \href {https://doi.org/10.18653/v1/2022.naacl-main.168} {{GPL}:
  Generative pseudo labeling for unsupervised domain adaptation of dense
  retrieval}.
\newblock In \emph{Proceedings of the 2022 Conference of the North American
  Chapter of the Association for Computational Linguistics: Human Language
  Technologies}, Seattle, United States. Association for Computational
  Linguistics.

\bibitem[{Wang et~al.(2013)Wang, Wang, Li, He, Chen, and Liu}]{ndcg}
Yining Wang, Liwei Wang, Yuanzhi Li, Di~He, Wei Chen, and Tie-Yan Liu. 2013.
\newblock A theoretical analysis of ndcg ranking measures.
\newblock In \emph{Proceedings of the 26th annual conference on learning theory
  (COLT 2013)}, volume~8, page~6. Citeseer.

\bibitem[{Wenzek et~al.(2020)Wenzek, Lachaux, Conneau, Chaudhary, Guzm{\'a}n,
  Joulin, and Grave}]{ccnet}
Guillaume Wenzek, Marie-Anne Lachaux, Alexis Conneau, Vishrav Chaudhary,
  Francisco Guzm{\'a}n, Armand Joulin, and Edouard Grave. 2020.
\newblock \href {https://aclanthology.org/2020.lrec-1.494} {{CCN}et: Extracting
  high quality monolingual datasets from web crawl data}.
\newblock In \emph{Proceedings of the 12th Language Resources and Evaluation
  Conference}, pages 4003--4012, Marseille, France. European Language Resources
  Association.

\bibitem[{Wolf et~al.(2020)Wolf, Debut, Sanh, Chaumond, Delangue, Moi, Cistac,
  Rault, Louf, Funtowicz, Davison, Shleifer, von Platen, Ma, Jernite, Plu, Xu,
  Le~Scao, Gugger, Drame, Lhoest, and Rush}]{transformers}
Thomas Wolf, Lysandre Debut, Victor Sanh, Julien Chaumond, Clement Delangue,
  Anthony Moi, Pierric Cistac, Tim Rault, Remi Louf, Morgan Funtowicz, Joe
  Davison, Sam Shleifer, Patrick von Platen, Clara Ma, Yacine Jernite, Julien
  Plu, Canwen Xu, Teven Le~Scao, Sylvain Gugger, Mariama Drame, Quentin Lhoest,
  and Alexander Rush. 2020.
\newblock \href {https://doi.org/10.18653/v1/2020.emnlp-demos.6} {Transformers:
  State-of-the-art natural language processing}.
\newblock In \emph{Proceedings of the 2020 Conference on Empirical Methods in
  Natural Language Processing: System Demonstrations}, pages 38--45, Online.
  Association for Computational Linguistics.

\bibitem[{Xin et~al.(2022)Xin, Xiong, Srinivasan, Sharma, Jose, and
  Bennett}]{modir}
Ji~Xin, Chenyan Xiong, Ashwin Srinivasan, Ankita Sharma, Damien Jose, and Paul
  Bennett. 2022.
\newblock \href {https://aclanthology.org/2022.findings-acl.316} {Zero-shot
  dense retrieval with momentum adversarial domain invariant representations}.
\newblock In \emph{Findings of the Association for Computational Linguistics:
  ACL 2022}, pages 4008--4020, Dublin, Ireland. Association for Computational
  Linguistics.

\bibitem[{Xin et~al.(2021)Xin, Xiong, Srinivasan, Sharma, Jose, and
  Bennett}]{xin2021zero}
Ji~Xin, Chenyan Xiong, Ashwin Srinivasan, Ankita Sharma, Damien Jose, and
  Paul~N Bennett. 2021.
\newblock Zero-shot dense retrieval with momentum adversarial domain invariant
  representations.
\newblock \emph{arXiv preprint arXiv:2110.07581}.

\bibitem[{Xiong et~al.(2020)Xiong, Xiong, Li, Tang, Liu, Bennett, Ahmed, and
  Overwijk}]{xiong2020approximate}
Lee Xiong, Chenyan Xiong, Ye~Li, Kwok-Fung Tang, Jialin Liu, Paul Bennett,
  Junaid Ahmed, and Arnold Overwijk. 2020.
\newblock Approximate nearest neighbor negative contrastive learning for dense
  text retrieval.
\newblock \emph{arXiv preprint arXiv:2007.00808}.

\bibitem[{Yang et~al.(2018)Yang, Qi, Zhang, Bengio, Cohen, Salakhutdinov, and
  Manning}]{yang2018hotpotqa}
Zhilin Yang, Peng Qi, Saizheng Zhang, Yoshua Bengio, William~W. Cohen, Ruslan
  Salakhutdinov, and Christopher~D. Manning. 2018.
\newblock {HotpotQA}: A {D}ataset for {D}iverse, {E}xplainable {M}ulti-hop
  {Q}uestion {A}nswering.
\newblock In \emph{Proceedings of the Conference on Empirical Methods in
  Natural Language Processing}, pages 2369--2380.

\bibitem[{Yu et~al.(2021)Yu, Xiong, and Callan}]{yu2021improving}
HongChien Yu, Chenyan Xiong, and Jamie Callan. 2021.
\newblock Improving query representations for dense retrieval with pseudo
  relevance feedback.
\newblock \emph{arXiv preprint arXiv:2108.13454}.

\bibitem[{Yu et~al.(2022)Yu, Xiong, Sun, Zhang, and Overwijk}]{yu2022coco}
Yue Yu, Chenyan Xiong, Si~Sun, Chao Zhang, and Arnold Overwijk. 2022.
\newblock Coco-dr: Combating distribution shifts in zero-shot dense retrieval
  with contrastive and distributionally robust learning.
\newblock \emph{arXiv preprint arXiv:2210.15212}.

\bibitem[{Zhang et~al.(2022)Zhang, Roller, Goyal, Artetxe, Chen, Chen, Dewan,
  Diab, Li, Lin et~al.}]{zhang2022opt}
Susan Zhang, Stephen Roller, Naman Goyal, Mikel Artetxe, Moya Chen, Shuohui
  Chen, Christopher Dewan, Mona Diab, Xian Li, Xi~Victoria Lin, et~al. 2022.
\newblock Opt: Open pre-trained transformer language models.
\newblock \emph{arXiv preprint arXiv:2205.01068}.

\bibitem[{Zhao et~al.(2021)Zhao, Xiong, Boyd-Graber, and
  Daum{\'e}~III}]{zhao2021distantly}
Chen Zhao, Chenyan Xiong, Jordan Boyd-Graber, and Hal Daum{\'e}~III. 2021.
\newblock \href {https://arxiv.org/abs/2110.04889} {Distantly-supervised
  evidence retrieval enables question answering without evidence annotation}.
\newblock \emph{arXiv preprint arXiv:2110.04889}.

\bibitem[{Zhong et~al.(2022)Zhong, Lei, and Chen}]{zhong2022training}
Zexuan Zhong, Tao Lei, and Danqi Chen. 2022.
\newblock Training language models with memory augmentation.
\newblock \emph{arXiv preprint arXiv:2205.12674}.

\end{thebibliography}
\bibliographystyle{acl_natbib}

\appendix
\clearpage
\section{Appendix}
\begin{table*}[h]
    \small
    \resizebox{\textwidth}{!}
    {\begin{tabular}{ l | l | l | c | c | c | c | c c c | c c }
        \toprule
         \multicolumn{1}{l}{\textbf{Split} ($\rightarrow$)} &
         \multicolumn{4}{c}{} &
         \multicolumn{1}{c}{\textbf{Train}}    &
         \multicolumn{1}{c}{\textbf{Dev}}    &
         \multicolumn{3}{c}{\textbf{Test}}   &
         \multicolumn{2}{c}{\textbf{Avg.~Word Lengths}} \\
         \cmidrule(lr){6-6}
         \cmidrule(lr){7-7}
         \cmidrule(lr){8-10}
         \cmidrule(lr){11-12}
           \textbf{Task ($\downarrow$)} &\textbf{Domain ($\downarrow$)} & \textbf{Dataset ($\downarrow$)} & \textbf{Title} & \textbf{Relevancy} & \textbf{\#Pairs} & \textbf{\#Query} & \textbf{\#Query} & \textbf{\#Corpus} & \textbf{Avg. D~/~Q } & \textbf{Query} & \textbf{Document} \\
         \midrule
    Passage-Retrieval    & Misc. & MS MARCO  & \xmark & Binary  & 532,761 &   ----  &   6,980   &   8,841,823      & 1.1 & 5.96  & 55.98  \\ \midrule[0.05pt] \midrule[0.05pt]
    Bio-Medical          & Bio-Medical & TREC-COVID   & \cmark & 3-level &   ----    &   ----  & 50     & 171,332   & 493.5& 10.60 & 160.77 \\
    Information          & Bio-Medical & NFCorpus    & \cmark & 3-level & 110,575 &  324  & 323    & 3,633     & 38.2 & 3.30  & 232.26 \\
    Retrieval (IR)       & Bio-Medical & BioASQ     & \cmark & Binary  & 32,916 & ---- & 500    & 14,914,602& 4.7  & 8.05  & 202.61 \\ \midrule
    Question             & Wikipedia  & NQ           & \cmark & Binary  & 132,803  &   ----  & 3,452 & 2,681,468 & 1.2  & 9.16  & 78.88  \\
    Answering       & Wikipedia  & HotpotQA     & \cmark & Binary  & 170,000 & 5,447 & 7,405  & 5,233,329 & 2.0  & 17.61 & 46.30  \\
     (QA)           &Finance& FiQA-2018   & \xmark & Binary  & 14,166  &  500  & 648    & 57,638    & 2.6  & 10.77 & 132.32 \\ \midrule
    Tweet-Retrieval      &Twitter& Signal-1M (RT)    & \xmark & 3-level &   ----    &   ----  & 97     & 2,866,316 & 19.6 & 9.30  & 13.93  \\ \midrule
    News      &News& TREC-NEWS      & \cmark & 5-level &   ----    &   ----  & 57     & 594,977 & 19.6 & 11.14  & 634.79  \\
    Retrieval      &News& Robust04  & \xmark & 3-level &   ----    &   ----  & 249   & 528,155 & 69.9 & 15.27  & 466.40  \\ \midrule
    Argument       & Misc. & ArguAna      & \cmark & Binary  &   ----    &   ----  & 1,406  & 8,674     & 1.0  & 192.98& 166.80 \\
    Retrieval   & Misc. & Touch\'e-2020 & \cmark & 3-level &   ----    &   ----  & 49     & 382,545   & 19.0 & 6.55  & 292.37 \\ \midrule
    Duplicate-Question   &StackEx.& CQADupStack   & \cmark & Binary  &   ----    &   ----  & 13,145 & 457,199   & 1.4  & 8.59  & 129.09 \\
    Retrieval            & Quora &  Quora        & \xmark & Binary  &   ----    & 5,000 & 10,000 & 522,931   & 1.6  & 9.53  & 11.44  \\ \midrule
    Entity-Retrieval     & Wikipedia  &  DBPedia      & \cmark & 3-level &   ----    &   67  & 400    & 4,635,922 & 38.2 & 5.39  & 49.68  \\ \midrule
    Citation-Prediction  & Scientific&  SCIDOCS       & \cmark & Binary  &   ----    &   ----  & 1,000  & 25,657    & 4.9  & 9.38  & 176.19 \\ \midrule
                         & Wikipedia  &  FEVER       & \cmark & Binary  & 140,085 & 6,666 & 6,666  & 5,416,568 & 1.2  & 8.13  & 84.76  \\ 
    Fact Checking        & Wikipedia  & Climate-FEVER  & \cmark & Binary  &   ----    &   ----  & 1,535  & 5,416,593 & 3.0  & 20.13 & 84.76  \\
                         & Scientific & SciFact     & \cmark & Binary  &   920      &   ----  &  300   & 5,183     & 1.1  & 12.37 & 213.63  \\
    \bottomrule
    \end{tabular}}
    \caption{Statistics of datasets in the BEIR benchmark. The table is taken from the original BEIR benchmark paper~\cite{beir}. 
    }
    \label{tab:dataset_stats}
\end{table*}

\subsection{Datasets Details}
\label{appx:datasets}
\paragraph{Evaluation Datasets}
Target domain datasets used in our experiments are collected in the BEIR benchmark~\citep{beir}\footnote{\url{https://github.com/beir-cellar/beir}} and include the following domains:
\begin{itemize}[leftmargin=*]
    \item Open-domain Question Answering (QA): HotpotQA~\citep{yang2018hotpotqa},  NQ~\citep{nq}, and FiQA~\citep{fiqa}.
    \item Bio-Medical Information Retrieval: TREC-COVID~\citep{treccovid}, NFCorpus~\citep{nfcorpus}, and BioASQ~\citep{bioasq}.
    \item Argument Retrieval: Webis-Touch\'e2020~\citep{touche} and ArguAna~\citep{arguana}.
    \item News Retrieval: TREC-NEWS~\citep{trecnews} and Robust04~\citep{robust04}.
    \item Tweet Retrieval: Signal-1m~\citep{signal1m}.
    \item Duplicate Question Retrieval: Quora~\citep{beir} and CQADupStack~\citep{cqadupstack}.
    \item Entity Retrieval: DBPedia~\citep{dbpedia}
    \item Citation Prediction:  SCIDOCS~\citep{scidocs}
    \item Fact Checking: SciFact~\citep{scifact}, FEVER~\citep{fever}, and Climate-FEVER~\citep{climatefever}
\end{itemize}
We list the statistics of the BEIR benchmark in Table~\ref{tab:dataset_stats}. 

\paragraph{Augmenting Corpora}
Corpus size
We first introduce more details on how we preprocessed the Medical Subject Headings (MeSH) Database.
We select text information from the Qualifier Record Set and Descriptor Record Set.
Each set contains multiple <Concept> elements, which is composed of three sub-elecments, i.e., <ConceptName>, <ScopeNote> and <TermList>.
Among the sub-elecments, <ScopeNote> is the major textual information source, which is usually a short description to a medical term or phenomenon.
We directly consider each <ScopeNote> as a document entry and concatenate it with corresponding <ConceptName>.

\begin{table}[t]
\centering
\small
\resizebox{0.85\textwidth}{!}{
\begin{tabular}{l c c}
\toprule
\textbf{Datasets} & \textbf{Corpus Size} & \textbf{Avg. Doc Length} \\
\hline
MS MARCO & 502,939 & 56.0 \\
MeSH & 32,326 & 16.8 \\
Wiki &21,015,324 &  100.0\\
\bottomrule
  \end{tabular}
}
\caption{Statistics of the augmenting corpora. \label{app-tab:dataset}} 
\end{table}

We list the statistics of the augmenting corpora in Table~\ref{app-tab:dataset}.

\subsection{Baselines}
\label{appx:baseline}
We use the baselines from the current BEIR leaderboard~\citep{beir} and recent papers. These baselines can be divided into four groups: dense retrieval,  dense retrieval with generated queries\footnote{We separate them from dense retrieval since they usually rely on Seq2seq models to generate pseudo query-document pairs, and they train a model for each dataset \emph{independently} instead of using a single model for all datasets.}, lexical retrieval and late interaction.
\paragraph{Dense Retrieval}
For dense retrieval, the baselines are the same dual-tower model as ours. We consider \textbf{DPR}~\citep{karpukhin2020dense}, \textbf{ANCE}~\citep{xiong2020approximate}, \textbf{T5-ANCE}, \textbf{coCondenser}~\cite{gao2022cocondenser} and one recently-proposed model \textbf{GTR}~\citep{gtr} with different size configuration in this paper.
\begin{itemize}[leftmargin=*]
    \item \textbf{DPR} uses a single BM25 retrieval example and in-batch examples as hard negative examples to train the model. Different from the original paper~\cite{beir} that train the DPR on QA datasets, we train DPR on MS MARCO~\cite{msmarco} Dataset for \emph{fair comparison}. Notice that this also lead to better results  according to \citet{modir}. 
    \item \textbf{ANCE}  constructs hard negative examples from an ANN index of the corpus. The hard negative training instances are updated in parallel during fine-tuning of the model. The model is a RoBERTa~\cite{liu2019roberta} model trained on MS MARCO for 600k steps.
     
    \item \textbf{T5-ANCE} Different with default ANCE setting, we replace the backbone language model RoBERTa with T5-base. 
    All the other model settings are the same with the original ANCE.
    We include this baseline because as a subroutine for \model{}, it could be viewed as an ablation without memory augmentation.
    We can directly observe the impact of plug-in mixture of memory by comparing T5-ANCE with \model{}.
    
    \item \textbf{coCondenser} is a continuous pre-trained model based on BERT, with the equivalent amount of parameters to BERT-base. It enhances the representation ability of \texttt{[CLS]} token by changing the connections between different layers of Transformer blocks. Fine-tuning of coCondenser uses BM25 and self-mined negatives.
    
    \item \textbf{Contriever} conducts unsupervised contrastive pretraining with data augmentations and momentum queues on Wikipedia and the larger CC-Net~\cite{ccnet} corpora for 500k steps.
     
     \item \textbf{GTR} initializes the dual encoders from the T5 models~\cite{raffel2019t5}. It is first pre-trained on
     Community QA\footnote{Unfortunately, this corpus has not been released by the authors.} with 2 billion question-answer pairs then fine-tuned on NQ and MS Marco dataset. In addition, they use the hard negatives released by RocketQA \cite{rocketqa} when finetuning with MS Marco data and the hard negatives release by \cite{lu-etal-2021-multi} for Natural Questions. 
     \textbf{GTR{\tiny{base}}} leverages the same T5-base model as \model{}, while \textbf{GTR{\tiny{large}}} is based on T5-large, which is not directly comparable to our method as it triples the parameters.
     
\end{itemize}

\paragraph{Dense Retrieval with Generated Queries} \textbf{GenQ} first fine-tunes a T5-base~\citep{raffel2019t5} model on MS MARCO for 2 epochs and then generate 5 queries for each passage as additional training data for the target domain to continue to fine-tune the TAS-B~\citep{tasb} model.

\paragraph{Lexical Retrieval}
Lexical retrieval is a score function for token matching calculated between two high-dimensional sparse vectors with token weights. 
\textbf{BM25}~\citep{bm25} is the most commonly used lexical retrieval function. We use the BM25 results reported in \citet{beir} for comparison.

\paragraph{Late Interaction}
We also consider a late interaction baseline, namely \textbf{ColBERT}~\citep{colbert}. The model computes multiple contextualized embeddings for each token of queries and documents, and then uses a maximum similarity function to retrieve relevant documents. This type of matching requires significantly more disk space for indexes and has a higher latency.

\subsection{Detailed Experimental Settings and hyperparameters}
\label{app:para}

Our implementation uses PyTorch~\citep{paszke2019pytorch} with Hugging Face Transformers~\citep{transformers}.
We optimize the model using AdamW~\citep{loshchilov2018adamw} with a peak learning rate at 5e-6, weight decay of 0.01, and linear learning rate decay. 
The global batch size is set to 256.
The maximum length of query and passage are set to 32 and 128 respectively.
We summarize all hyperparameter settings in Table~\ref{app-tab:hyper_setting}.
The model is trained with 8 Nvidia A100 80GB GPUs and FP16 mixed-precision training. 
The total running time is 6.6 hrs for three episodes of augmentation component training and 6.3 hrs for end retriever training.
We detail the training time of each episode in Table~\ref{app-tab:train-time}.

When evaluating on the BEIR benchmark, we follow the setting in GTR~\citep{gtr}, which use sequences of 64 tokens for the questions and 512 for the documents in all datasets except Trec-News, Robust-04 and ArguAna. In particular, we set the document length to 768 for Trec-News and Robust-04.
For ArguAna, we set both question and document length to 128.
The above length setting is in accordance to the average query and document lengths in these datasets.

\begin{table}[t]
\centering
\small
\resizebox{0.9\textwidth}{!}{
\begin{tabular}{l l}
\toprule
\textbf{Hyperparameters} & \textbf{Settings} \\
\hline
Grounding document number & 10 \\
Attention threshold number & 5 \\
Negative mining depth & 200 \\
Global batch size (query size per batch) & 256 \\
Positive number per query & 1 \\
Negative number per query & 7 \\
Peak learnig rate & 5e-6 \\
Learnig rate decay & 0.01 \\
Optimizer & AdamW \\
Scheduler & Linear \\
MARCO Maximum query length & 32 \\
MARCO Maximum document length & 128 \\

\bottomrule
  \end{tabular}
}
\caption{The hyperparameters of \model{}.\label{app-tab:hyper_setting}} 
\end{table}

\begin{table}[t]
\centering
\small
\resizebox{\textwidth}{!}{
\begin{tabular}{lcc}
\toprule
\textbf{Stage} & \textbf{Augmentation Component}  & \textbf{End Retriever}\\
\hline
Epi-1 & 0.8h& 1.5h \\
Epi-2 & 0.8h& 1.5h \\
Epi-3 & 0.8h& 1.5h \\
Index refresh & 1.4h& 0.6h \\
Refresh number & 3 & 3 \\
\hline
Overall & 6.6h & 6.3h \\
\bottomrule
  \end{tabular}
}
  \caption{Training time for \model{} with three training episodes. We use 8 Nvidia A100 80GB GPUs with FP16 mixed-precision training.\label{app-tab:train-time}}
\end{table}



\end{document}